\documentclass[10pt,journal,compsoc]{IEEEtran}
\ifCLASSOPTIONcompsoc
  \usepackage[nocompress]{cite}
\else
  \usepackage{cite}
\fi
\ifCLASSINFOpdf
  \usepackage[pdftex]{graphicx}
\else
\fi
\usepackage[tbtags]{amsmath}
\usepackage{dblfloatfix}
\usepackage{etoolbox}

\usepackage{amssymb}
\usepackage{bm}

\usepackage[hidelinks]{hyperref}

\usepackage[theorems]{tcolorbox}
\tcbset{
    colback=white,
    colframe=black,
    boxrule=0.75pt
}
\tcbset{highlight math style={
    colback=white,
    colframe=black,
    boxrule=0.75pt
}}

\usepackage{xspace}
\newcommand{\ie}{i.e.\xspace}
\newcommand{\eg}{e.g.\xspace}
\newcommand{\etal}{\textit{et~al.\xspace}}

\usepackage{booktabs}
\usepackage{tabularx}
\newcolumntype{L}[1]{>{\raggedright\arraybackslash}m{#1}}
\newcolumntype{C}[1]{>{\centering\arraybackslash}m{#1}}
\newcolumntype{R}[1]{>{\raggedleft\arraybackslash}m{#1}}
\usepackage{multirow}

\usepackage{siunitx}

\newcommand{\abs}[1]{\left\vert#1\right\vert}
\newcommand{\norm}[1]{\left\Vert#1\right\Vert}
\newcommand{\dotproduct}[2]{\left\langle#1\middle\vert#2\right\rangle}
\newcommand{\transpose}{\mkern-1.5mu\mathsf{T}}

\newcommand{\smallabs}[1]{\vert#1\vert}
\newcommand{\smallnorm}[1]{\Vert#1\Vert}

\renewcommand{\vec}[1]{\bm{#1}}
\newcommand{\mat}[1]{\bm{#1}}
\newcommand{\set}[1]{\mathcal{#1}}
\newcommand{\given}{\mkern 2mu \vert \mkern 2mu}
\newcommand{\tightsetminus}{\mkern 2mu {\setminus} \mkern 2mu}

\DeclareMathOperator{\erf}{erf}
\DeclareMathOperator{\erfc}{erfc}
\DeclareMathOperator{\hyp1f1}{{\mkern -1mu}_1F_1}
\DeclareMathOperator{\bessel}{J}

\newcommand{\euler}{\mathrm{e}}
\newcommand{\imag}{\mathrm{i}}

\newcommand{\identity}{\mathrm{I}}

\let\originalleft\left
\let\originalright\right
\renewcommand{\left}{\mathopen{}\mathclose\bgroup\originalleft}
\renewcommand{\right}{\aftergroup\egroup\originalright}

\begin{document}
\title{Incomplete Gamma Kernels: Generalizing Locally Optimal Projection Operators}

\author{Patrick~Stotko,
        Michael~Weinmann,
        and~Reinhard~Klein% <-this % stops a space
\IEEEcompsocitemizethanks{\IEEEcompsocthanksitem P. Stotko and R. Klein are with the Institute of Computer Science II -- Visual Computing, University of Bonn, Germany.\protect\\
E-mail: stotko@cs.uni-bonn.de, rk@cs.uni-bonn.de
\IEEEcompsocthanksitem M. Weinmann is with the Department of Intelligent Systems, Delft University of Technology, The Netherlands\protect\\
E-mail: m.weinmann@tudelft.nl}% <-this % stops an unwanted space
\thanks{Manuscript received MM DD, 20YY; revised MM DD, 20YY.}}


\markboth{IEEE Transactions on Pattern Analysis and Machine Intelligence}%
{Stotko \MakeLowercase{\textit{et al.}}: Incomplete Gamma Kernels: Generalizing Locally Optimal Projection Operators}



\IEEEtitleabstractindextext{%
\begin{abstract}
We present incomplete gamma kernels, a generalization of Locally Optimal Projection (LOP) operators.
In particular, we reveal the relation of the classical localized $ L_1 $ estimator, used in the LOP operator for point cloud denoising, to the common Mean Shift framework via a novel kernel.
Furthermore, we generalize this result to a whole family of kernels that are built upon the incomplete gamma function and each represents a localized $ L_p $ estimator.
By deriving various properties of the kernel family concerning distributional, Mean Shift induced, and other aspects such as strict positive definiteness, we obtain a deeper understanding of the operator's projection behavior.
From these theoretical insights, we illustrate several applications ranging from an improved Weighted LOP (WLOP) density weighting scheme and a more accurate Continuous LOP (CLOP) kernel approximation to the definition of a novel set of robust loss functions.
These incomplete gamma losses include the Gaussian and LOP loss as special cases and can be applied to various tasks including normal filtering.
Furthermore, we show that the novel kernels can be included as priors into neural networks.
We demonstrate the effects of each application in a range of quantitative and qualitative experiments that highlight the benefits induced by our modifications.
\end{abstract}

}

\maketitle

\IEEEdisplaynontitleabstractindextext

% \fi
\IEEEpeerreviewmaketitle

%-------------------------------------------------------------------------
\IEEEraisesectionheading{\section{Introduction}\label{sec:introduction}}


%\IEEEpubidadjcol

\begin{figure*}
    \centering
    \includegraphics[width=\linewidth, height=0.25\textheight, keepaspectratio]{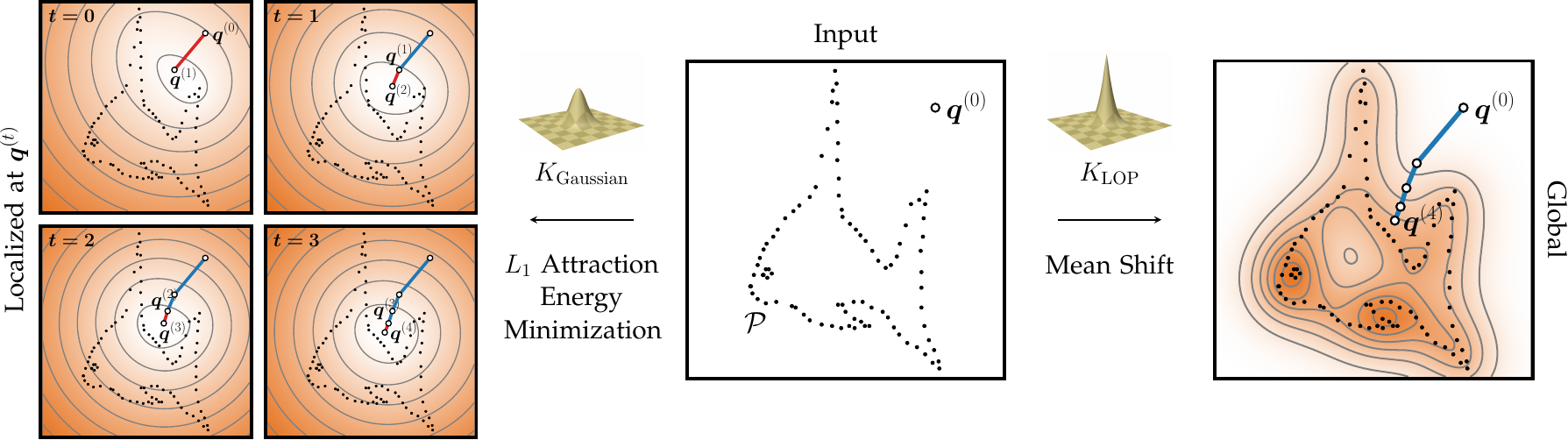}
    \caption{Relation between LOP and Mean Shift in the example of the \emph{2D Fish} model. Minimizing the \emph{localized} $ L_1 $ attraction energy with the Gaussian kernel $ K_{\mathrm{Gaussian}} $ (\emph{left}) results in the same trajectory $ \vec{q}^{(t)} $ as applying Mean Shift on a \emph{global} kernel density estimate with the kernel $ K_{\mathrm{LOP}} $ (\emph{right}).}
    \label{fig:equivalence}
\end{figure*}

\IEEEPARstart{D}{igital} 3D scene models have become a crucial prerequisite for numerous applications in entertainment, advertisement, design, architecture, autonomous systems, and cultural heritage.
In this context, the accurate digitization of real-world objects and scenes is of great relevance and offers new opportunities regarding a variety of tasks including AR/VR-based inspection and collecting realistic training data for tasks in robotics, autonomous driving, aerial or satellite surveys.
Aside from professional scanning campaigns with expensive laser scanning equipment, there has also been an increasing trend towards more practical scene capture with consumer-grade hardware such as passive purely image-based scene scanning using Structure-from-Motion and Multi-view Stereo approaches, or with respective cheaper active time-of-flight depth sensors that have meanwhile even been integrated into numerous mobile devices.
However, the use of passive scene scanning or active scanning based on cheap hardware with low sensor quality and low sensor resolution induces noise in the capture process and thereby results in noisy point clouds and a low number of points that might not preserve finer geometric details, which, in turn, may lead to artifacts in the registration and subsequent surface reconstruction procedures.
Furthermore, the limited accessibility of capture conditions as well as occlusions induce holes and highly irregular samplings and distributions of the captured data.
These challenges result in an increasing interest in robust filtering techniques that are capable of handling noise, outliers, registration artifacts, as well as irregularly-sampled and missing data and can provide a clean, denoised, and uniformly resampled point cloud suitable for high-fidelity surface reconstruction.

Among others, the Locally Optimal Projection (LOP) operator~\cite{lipman2007parameterization} has gained a lot of attention in recent years due to its benefit of not relying on a well-defined surface parametrization or a piecewise planar approximation and, meanwhile, there has been a whole series of further extensions of this approach~\cite{huang2009consolidation,liao2013efficient,huang2013edge,preiner2014continuous}.
Furthermore, many learning-based approaches also aim at projecting the noisy data onto a (latent) denoised manifold~\cite{zhang2020pointfilter,xu2022tdnet}.
Therefore, investigations towards the unification of traditional approaches with their respective regularization techniques might be of great relevance for future learning-based approaches as well.
Even further, traditional techniques, and particularly those approaches that incorporate probabilistic modeling of the data such as the density-based Mean Shift clustering method~\cite{fukunaga1975estimation,cheng1995mean,comaniciu2002mean}, become more and more relevant in modern deep learning methods.
Besides their application to structure the latent space representation of the data within encoder-decoder approaches~\cite{madaan2019deep}, there is even a direct relation between the Mean Shift approach and denoising autoencoders~\cite{arjomand2018image}.
In particular, as the output of an optimal denoising autoencoder corresponds to the local mean of the true data density~\cite{alain2014regularized}, the autoencoder loss can be interpreted as a Mean Shift vector~\cite{arjomand2018image}.
However, to the best of our knowledge, this observation has not yet been explored in the context of point cloud denoising.
Hence, relating traditional concepts to modern deep learning methods might not only lead to a more explainable behavior of the latter but also allow increasing the resulting performance.
In turn, this relies on the better understanding of the relationship between previous (traditional) techniques.

In this paper, we investigate the theoretical relationship of projection-based point cloud denoising approaches with their respective properties and show that these are unified within the common probabilistic Mean Shift framework.
In particular, the key contributions of our work are:
\begin{itemize}
  \item We reveal the relation of the classical localized $ L_1 $ estimator used in LOP to the Mean Shift framework via a novel kernel $ K_{\mathrm{LOP}} $ and introduce the family of \emph{incomplete gamma kernels} $ K_{\Gamma} $ as a generalization of this result where each kernel represents a localized $ L_p $ estimator (see Section~\ref{sec:background}).
  \item We derive various properties of the kernel family concerning distributional, Mean Shift induced, and other aspects such as strict positive definiteness to obtain a deeper understanding of the operator's projection behavior (see Section~\ref{sec:kernel_properties}).
  \item We demonstrate that leveraging the derived theoretical insights enables several applications including an improved Weighted LOP (WLOP) density weighting scheme, a more accurate Continuous LOP (CLOP) kernel approximation, the derivation of \emph{incomplete gamma losses}, a set of novel robust loss functions, as well as neural network priors (see Section~\ref{sec:applications}).
\end{itemize}
In our evaluation, we demonstrate the benefits induced by our modifications in a range of quantitative and qualitative experiments.
Furthermore, the theoretical insights of our investigations with their proven effect may be of great relevance also for future learning-based approaches.
The source code is available at \url{https://github.com/stotko/incomplete-gamma-kernels}.

%-------------------------------------------------------------------------
\section{Related Work}
\label{sec:related_work}

In the following, we provide a review of geometric and learning-based denoising approaches.
Furthermore, we also review seminal work regarding the theory and application of the Mean Shift framework due to its relationship to LOP approaches that we will demonstrate later.

\subsection{Geometric Denoising Approaches}

Following early approaches such as the local fitting of tangent planes~\cite{hoppe1992surface} or using radial basis functions~\cite{carr2001reconstruction}, respective developments particularly focused on projection-based methods, sparsity-based methods and non-local methods.

\textbf{Projection-based Methods.}
These approaches rely on the assumption of an underlying smooth surface and the projection of noisy data points onto the estimated local surface.
For this purpose, respective approaches apply moving least squares (MLS) based methods~\cite{alexa2003computing,amenta2004defining,fleishman2005robust,oztireli2009feature}, robust principal component analysis (RPCA)~\cite{narvaez2006point} and moving robust principal component analysis (MRPCA)~\cite{mattei2017point}, or locally optimal projection based operators where the LOP operator~\cite{lipman2007parameterization} has been extended in terms of Weighted LOP (WLOP)~\cite{huang2009consolidation}, Feature LOP (FLOP)~\cite{liao2013efficient}, Continuous LOP (CLOP)~\cite{preiner2014continuous}, Edge-Aware Resampling (EAR)~\cite{huang2013edge} and a Gaussian mixture model inspired projection operator~\cite{lu2017gpf}.
The latter has been demonstrated to be capable of resampling point clouds while preserving features due to the additional guidance of filtered normals.

\textbf{Sparsity-based Methods.}
This class of approaches relies on the assumption that objects can be represented in terms of piecewise smooth surfaces with sparse features.
Respective denoising techniques include $ L_0 $\mbox{-}norm~\cite{sun2015denoising,cheng2019efficient} and $ L_1 $\mbox{-}norm minimization~\cite{avron2010l1,mattei2017point,leal2020sparse}, sparse dictionary learning~\cite{digne2017sparse} as well as patch-based or feature-based graph Laplacian regularization~\cite{zeng20193d,hu2020feature,dinesh2020point}, graph-based point cloud denoising based on jointly leveraging geometry and color information~\cite{irfan2021joint}, guided filtering based on normal information followed by a $ L_1 $\mbox{-}medial skeleton extraction to get the sharp structure of the surface~\cite{zheng2017guided} as well as leveraging gravitational feature functions~\cite{shi2022three}.
In the context of denoising dynamic point clouds, Hu \etal~\cite{hu2021dynamic} explored the temporal coherence of spatio-temporal graphs with respect to the underlying surface, where a respective manifold-to-manifold distance has been introduced.
Furthermore, data-driven exemplar priors have been used for surface reconstruction~\cite{remil2017surface}, where the sparsity of local shapes from a collection of 3D objects has been explored.

\textbf{Non-local Methods.}
In contrast to the previous classes, these approaches rely on the assumption that geometric statistics are (approximately) shared by certain surface patches of a 3D model, \ie local surface denoising is conducted based on collected neighborhoods with similar geometry~\cite{rosman2013patch,lu2020low,chen2019multi,zhu2022non}.
However, the definition of a suitable metric as well as the regular representation of local surface structures remain challenging.
Furthermore, density-based point cloud denoising has been approached by first applying particle-swarm based optimization for kernel density estimation followed by a Mean Shift clustering-based outlier removal and a final bilateral mesh filtering~\cite{zaman2017density}.

\subsection{Learning-based Denoising Approaches}

Recent works more and more leverage deep learning for point cloud denoising as well as surface reconstruction from point clouds.
Examples include approaches for point cloud consolidation and resampling such as \mbox{PointNet}~\cite{qi2017pointnet}, \mbox{PointNet++}~\cite{qi2017pointnet++}, patch-based progressive point cloud upsampling~\cite{yifan2019patch} as well as the unification of the considerations of densifying, denoising and completing point clouds~\cite{choe5194deep}.
Other approaches followed the principles of initially projecting the points onto coarse-level local reference planes and applying a subsequent refinement~\cite{duan20193d} or the initial removal of outliers before conducting the denoising~\cite{rakotosaona2020pointcleannet}.
Further approaches include edge-aware point cloud consolidation~\cite{yu2018ec}, adversarial defense~\cite{zhou2019dup}, graph-convolutional methods~\cite{pistilli2020learning}, unsupervised approaches such as Total Denoising~\cite{hermosilla2019total}, gradient field based denoising~\cite{luo2021score,chen2021deep,zhao2022point}, differentiable approaches~\cite{roveri2018pointpronets,yifan2019differentiable,luo2020differentiable} as well as manifold learning based on encoder-decoder architectures~\cite{zhang2020pointfilter,xu2022tdnet}.
Non-local self-similarities have also been considered to define neural self-priors that capture geometric repetitions~\cite{hanocka2020point2mesh}, capture semantically related non-local features~\cite{huang2020non}, or apply self-correction by allowing the model to capture structural and contextual information from initially disorganized parts~\cite{chen2021shape}.
Furthermore, normalizing flows have been applied to the learn the distribution of noisy points and disentangle noise from the latent space~\cite{mao2022pd}.
In addition, the feature-aware recurrent point cloud denoising network (\mbox{RePCD-Net})~\cite{chen2022repcd} combines a recurrent network architecture for noise removal with multi-scale feature aggregation and propagation and a feature-aware Chamfer distance loss.

\subsection{Mean Shift Approaches}

The Mean Shift approach~\cite{fukunaga1975estimation} is a well-studied local mode-seeking method with diverse applications including data clustering~\cite{cheng1995mean,comaniciu2002mean,grillenzoni2016design,beck2019distributed}, image filtering~\cite{comaniciu2002mean}, segmentation~\cite{comaniciu2002mean,jang2021meanshift++}, denoising~\cite{arjomand2017deep,arjomand2018image}, and object tracking~\cite{jang2021meanshift++}.
Tremendous effort has been spent to study its convergence behavior~\cite{cheng1995mean,comaniciu2002mean,li2007note,chen2015convergence,ghassabeh2015sufficient,huang2018convergence} which culminated in a rigorous set of properties proven by Yamasaki and Tanaka~\cite{yamasaki2019properties}.
Recently, Mean Shift clustering has also been applied in the latent space of neural encoder-decoder approaches to achieve a better structured data representation~\cite{madaan2019deep}.
Furthermore, the connection between the Mean Shift approach and denoising autoencoders~\cite{vincent2008extracting} has been revealed by Bigdeli \etal~\cite{arjomand2018image}, who leveraged the observation that the output of an optimal denoising autoencoder (DAE) is a local mean of the true data density~\cite{alain2014regularized} to show that that the autoencoder loss is a Mean Shift vector and to use the respective magnitude to define a prior for image restoration.

%-------------------------------------------------------------------------
\section{Background}
\label{sec:background}

\begin{figure*}[b]
    \centering
    \includegraphics[width=\linewidth, height=0.2\textheight, keepaspectratio]{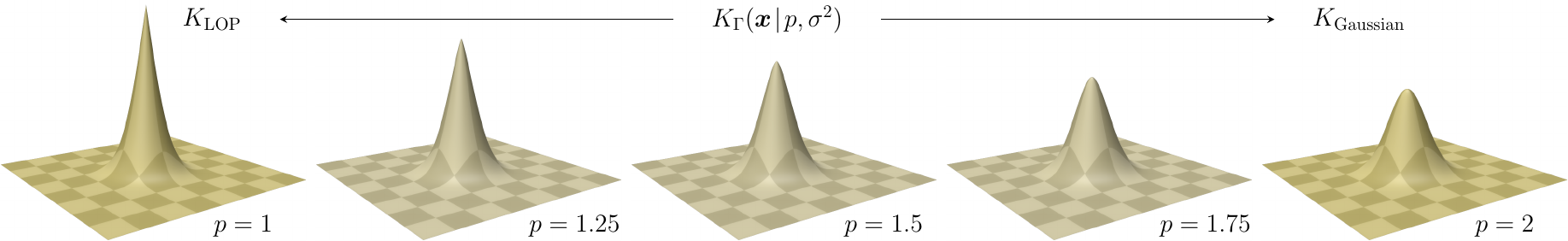}
    \caption{Interpolation between 2D incomplete gamma kernels $ K_{\Gamma} $ with varying $ { p \in [1, 2] } $ and fixed $ { \sigma^2 = 1 / 32 } $. Each kernel corresponds to a localized attraction energy minimization with the respective $ p $\mbox{-}norm.}
    \label{fig:gamma_kernels}
\end{figure*}

Before deriving our proposed kernel family as a generalization of LOP in the context of Mean Shift, we first provide a brief introduction into the concepts of both approaches.

\subsection{Mean Shift}
\label{sec:mean_shift}

The basic objective of Mean Shift~\cite{fukunaga1975estimation} is to find the modes of a probabilistic distribution $ f $ which has been observed by a (sparse) set of points $ { \set{P} = \{ \vec{p}_i \in \mathbb{R}^d \} } $ in a $ d $\mbox{-}dimensional vector space.
Due to this abstract formulation, it has been applied to various computer vision problems by choosing an appropriate application-specific feature space, \eg the $ L^{*} u^{*} v^{*} $ color space for image filtering or segmentation \cite{comaniciu2002mean}.
In order to analyze the unknown distribution $ f $ at any point $ \vec{q} \in \mathbb{R}^d $, it is modeled by a kernel density estimate:
\begin{equation}
    \hat{f}_{\set{P},K}(\vec{q})
    =
    \frac{1}{\abs{\set{P}} \, h^d} \sum_i {\textstyle K( \frac{\vec{p}_i - \vec{q}}{h}) }
    \label{eq:kernel_density_estimate}
\end{equation}
Here, $ h $ denotes the kernel window size and $ K $ a kernel that is non-negative ($ { K(\vec{x}) \ge 0 } $), normalized ($ { \int_{\mathbb{R}^d} K(\vec{x}) \, \mathrm{d} \vec{x} = 1 } $), and radially symmetric ($ { K(\vec{x}) = c_K \, k(\norm{\vec{x}}^2) } $).
The 1\mbox{-}dimensional function $ k $ defined in the symmetry constraint, where $ c_K $ denotes a normalization constant, is called the kernel profile of $ K $ and plays an important role in the analysis of Mean Shift~\cite{yamasaki2019properties}.
Furthermore, the gradient of the kernel density estimate
\begin{equation}
    \nabla \hat{f}_{\set{P},K}(\vec{q})
    =
    \frac{2}{\abs{\set{P}} \, h^{d + 2}} \frac{c_K}{c_G} \sum_i {\textstyle G( \frac{\vec{p}_i - \vec{q}}{h}) } \, (\vec{p}_i - \vec{q})
    \label{eq:kernel_density_estimate_gradient}
\end{equation}
can be derived using the kernel $ { G(\vec{x}) = c_G \, g(\norm{\vec{x}}^2) } $ with the normalization constant $ c_G $ and its corresponding profile $ { g(x) = - \frac{\mathrm{d}}{\mathrm{d} x} k(x) } $.
Based on these two functions, the main component of the algorithm for finding the modes is the Mean Shift vector~\cite{comaniciu2002mean}
\begin{equation}
    \vec{m}_{\set{P},G}(\vec{q})
    =
    \frac{h^2}{2} \frac{c_G}{c_K} \frac{\nabla \hat{f}_{\set{P},K}(\vec{q})}{\hat{f}_{\set{P},G}(\vec{q})}
    =
    \frac{\sum_i G( \frac{\vec{p}_i - \vec{q}}{h}) \, (\vec{p}_i - \vec{q})}{\sum_i G( \frac{\vec{p}_i - \vec{q}}{h}) \, \phantom{(\vec{p}_i - \vec{q})}}
\end{equation}
which describes the gradient vector normalized with respect to the kernel $ G $.
In particular, it directly determines the updated point at each time step $ t $ by the corresponding fixed-point iteration $ { \vec{q}^{(t + 1)} = \vec{q}^{(t)} + \vec{m}_{\set{P},G}(\vec{q}^{(t)}) } $ which performs gradient ascent on the kernel density estimate $ \hat{f}_{\set{P},K} $.

\subsection{Locally Optimal Projection}
\label{sec:locally_optimal_projections}

{Unlike approaches that are based on the probabilistic concept of Mean Shift to tackle various different computer vision tasks, the problem of denoising point clouds has also been of great relevance for the computer graphics community.
Many advanced solutions have been independently developed there, especially methods leveraging the robust $ L_1 $ median.
To efficiently compute the unique global solution in these median-based formulations, the iterative Weiszfeld algorithm~\cite{weiszfeld1937point} has become a popular and commonly used choice.
Specifically, it also marks the foundation of LOP~\cite{lipman2007parameterization} as a robust localized 3D projection operator.
Given a set of noisy 3D target points $ { \set{P} = \{ \vec{p}_i \in \mathbb{R}^3 \} } $ sampled from a smooth surface $ \set{S} $, the task here consists in projecting and uniformly distributing an independent set of 3D points $ { \set{Q} = \{ \vec{q}_j \in \mathbb{R}^3 \} } $ onto the unknown surface $ \set{S} $ which is defined by the observations $ \set{P} $ only.
This can be expressed in terms of an energy formulation
\begin{equation}
    E(\set{Q})
    =
    \sum_j E_{\mathrm{LOP}}(\vec{q}_j) + E_{\mathrm{rep}}(\vec{q}_j)
\end{equation}
based on an attraction and a repulsion term
\begin{gather}
    E_{\mathrm{LOP}}(\vec{q}_j)
    =
    \sum_i \theta(\smallnorm{\vec{p}_i - \vec{q}_j^{(t)}}) \, \smallnorm{\vec{p}_i - \vec{q}_j}
    \label{eq:attraction_energy}
    \\
    E_{\mathrm{rep}}(\vec{q}_j)
    =
    \lambda_j \sum_{i, i \not = j} \theta(\smallnorm{\vec{q}_i^{(t)} - \vec{q}_j^{(t)}}) \, \eta(\smallnorm{\vec{q}_i^{(t)} - \vec{q}_j})
    \label{eq:repulsion_energy}
\end{gather}
where $ { \theta(x) = \euler^{- x^2 / (h / 4)^2} } $ denotes a compact localization kernel and $ \eta $ a decreasing regularization function penalizing small distances between projection points to ensure a uniform distribution of the projected points.
Common choices of $ \eta $ include the originally proposed function $ { \eta_{\mathrm{LOP}}(x) = 1 / (3 x^3) } $~\cite{lipman2007parameterization} as well as the less rapidly decreasing function $ { \eta_{\mathrm{WLOP}}(x) = - x } $~\cite{huang2009consolidation}.
Both energy terms are balanced by weights $ \lambda_j $ which are chosen such that they only depend on a single, global parameter $ { \mu \in [0, 1 / 2) } $.
Based on the Weiszfeld algorithm, the solution to this optimization problem can be obtained by the fixed-point iteration
\begin{equation}
\begin{split}
    \vec{q}_j^{(t + 1)}
    = \ &
    \frac{\sum_i \alpha(\smallnorm{\vec{p}_i - \vec{q}_j^{(t)}}) \, \vec{p}_i}{\sum_i \alpha(\smallnorm{\vec{p}_i - \vec{q}_j^{(t)}}) \, \phantom{\vec{p}_i}}
    \\
    & +
    \mu \, \frac{\sum_{i, i \not = j} \beta(\smallnorm{\vec{q}_i^{(t)} - \vec{q}_j^{(t)}}) \, (\vec{q}_j^{(t)} - \vec{q}_i^{(t)})}{\sum_{i, i \not = j} \beta(\smallnorm{\vec{q}_i^{(t)} - \vec{q}_j^{(t)}}) \, \phantom{(\vec{q}_j^{(t)} - \vec{q}_i^{(t)})}}
    \label{eq:lop_update}
\end{split}
\end{equation}
with kernels $ { \alpha(x) = \theta(x) /x } $ and $ { \beta(x) = \theta(x) / x \, \abs{\frac{\mathrm{d}}{\mathrm{d} x} \eta(x)} } $.

\subsection{Generalization via Incomplete Gamma Kernels}
\label{sec:generalization_via_incomplete_gamma_kernels}

Although Mean Shift and Locally Optimal Projection have been separately developed from different contexts and mathematical concepts with the aim of solving a distinct problem, we can link both approaches by rewriting the update step in \eqref{eq:lop_update}:
\begin{equation}
    \vec{q}_j^{(t + 1)}
    =
    \vec{q}_j^{(t)} + \vec{m}_{\set{P},G_{\mathrm{LOP}}}(\vec{q}_j^{(t)}) - \mu \, \vec{m}_{\set{Q}_j^{(t)},G_{\mathrm{rep}}}(\vec{q}_j^{(t)})
\end{equation}
This reveals that the LOP operator is a combination of two Mean Shift steps in the 3D space:
1) a standard Mean Shift with respect to the target set $ \set{P} $ and $ G_{\mathrm{LOP}} $ being the normalized kernel $ \alpha $;
and 2) a reverse applied Blurring Mean Shift~\cite{cheng1995mean} where the Mean Shift vector is instead subtracted and computed from the shifted and, in turn, blurred source set $ { \set{Q}_j^{(t)} = \set{Q}^{(t)} \tightsetminus \{ \vec{q}_j^{(t)} \} } $ as well as from the kernel $ G_{\mathrm{rep}} $ corresponding to the normalized kernel $ \beta $.
Therefore, we can interpret the localized $ L_1 $ attraction energy minimization with a Gaussian kernel in \eqref{eq:attraction_energy} as a maximization of a global kernel density estimate \eqref{eq:kernel_density_estimate} with respect to a different kernel $ K_{\mathrm{LOP}} $.
An example of this relation is shown in Fig.~\ref{fig:equivalence}.

To derive $ K_{\mathrm{LOP}} $ as part of a novel kernel family $ K_{\Gamma} $ for the general case of the $ \mathbb{R}^d $ space, we consider the 1\mbox{-}dimensional profile of the involved kernel $ G_{\mathrm{LOP}} $, \ie $ \alpha $ in \eqref{eq:lop_update}, which resembles a gamma distribution $ { f_{\Gamma}(x \given a, b) \propto x^{a - 1} \, \euler^{- x / b} } $ with support $ { x \in (0, \infty) } $ and parameters $ { a > 0, b > 0 } $.
The profile of the actual kernel then resembles the distribution $ { \bar{F}_{\Gamma}(x \given a, b) \propto \Gamma(a, x / b) } $ which is the complementary CDF of $ f_{\Gamma} $ and based on the \emph{upper incomplete gamma function} $ { \Gamma(a, x) = \int_{x}^{\infty} t^{a - 1} \, \euler^{-t} \, \mathrm{d} t } $.
Therefore, the $ d $\mbox{-}dimensional kernel has the general form $ { K_{\Gamma}(\vec{x} \given a, b) = c_{K_{\Gamma}} \, \Gamma(a, \norm{\vec{x}}^2 / b) } $.

Since we also need to compute the respective normalization constant $ c_{K_{\Gamma}} $, we switch the integration domain to spherical coordinates and substitute $ { s = r^2 / b } $:
\begin{equation}
\begin{split}
    \frac{1}{c_{K_{\Gamma}}}
    & =
    \int_{\mathbb{R}^d} {\textstyle \Gamma(a, \frac{\norm{\vec{x}}^2}{b}) } \, \mathrm{d} \vec{x}
    =
    \int_{\Omega} \int_{0}^{\infty} {\textstyle \Gamma(a, \frac{r^2}{b}) } \, r^{d - 1} \, \mathrm{d} r \, \mathrm{d} \vec{\Omega}
    \\
    & =
    \frac{b^{\frac{d}{2}}}{2} \left[ \int_{\Omega} \, \mathrm{d} \vec{\Omega} \right] \, \left[ \int_{0}^{\infty} {\textstyle \Gamma(a, s) } \, s^{\frac{d}{2} - 1} \, \mathrm{d} s \right]
\end{split}
\end{equation}
Due to radial symmetry, both integrals can be solved independently.
The former one describes the surface area of the $ d $\mbox{-}dimensional unit sphere $ \Omega $ and has the closed form $ { \int_{\Omega} \, \mathrm{d} \vec{\Omega} = 2 \pi^{d / 2} / \Gamma(d / 2) } $.
Using the relation $ { \int_{0}^{\infty} \Gamma(a, x) \, x^{b - 1} \, \mathrm{d} x = \Gamma(a + b) / b } $~\cite{bateman1953higher}, we get an expression for the latter one in terms of the ordinary gamma function.
We can also apply the recursive relation of the gamma function $ { \Gamma(a + 1) = a \, \Gamma(a) } $ and conclude that $ { 1 / c_{K_{\Gamma}} = (\pi b)^{d / 2} \, \Gamma(d / 2 + a) / \Gamma(d / 2 + 1) } $.

Finally, we change the parametrization by setting $ { a = p / 2, b = 2 \sigma^2 } $ to obtain the final kernel:
%\begin{tcolorbox}[ams align]
\begin{equation}
\tcbhighmath{
    K_{\Gamma}(\vec{x} \given p, \sigma^2)
    =
    \frac{1}{(2 \pi \sigma^2)^{\frac{d}{2}}} \frac{\Gamma(\frac{d + 2}{2})}{\Gamma(\frac{d + p}{2})} \, {\textstyle \Gamma(\frac{p}{2}, \frac{\norm{\vec{x}}^2}{2 \sigma^2}) }
}
\end{equation}
%\end{tcolorbox}
These \emph{incomplete gamma kernels} span a family of Mean Shift kernels corresponding to $ L_p $ estimators of the attraction energy localized by a Gaussian kernel.
An important special case of this family is the LOP kernel for which we choose $ { p = 1, \sigma^2 = 1 / 32 } $ and apply the identity $ { \Gamma(1 / 2, x) = \sqrt{\pi} \, \erfc(\sqrt{x}) } $ to get
\begin{equation}
    K_{\mathrm{LOP}}(\vec{x})
    =
    \frac{4^d}{\pi^{\frac{d - 1}{2}}} \frac{\Gamma(\frac{d + 2}{2})}{\Gamma(\frac{d + 1}{2})} \, \erfc(4 \, \norm{\vec{x}})
    \label{eq:lop_kernel}
\end{equation}
where $ \erfc $ denotes the \emph{complementary error function}.
Another special case is the corresponding Gaussian kernel $ K_{\mathrm{Gaussian}} $ obtained by setting $ { p = 2 } $ which is a common choice in Mean Shift and has been extensively analyzed as the localized $ L_2 $ estimator of the geometric mean.
Fig.~\ref{fig:gamma_kernels} shows an interpolation between these kernels by varying the $ p $\mbox{-}norm.

\begin{table}
    \renewcommand{\arraystretch}{1.25}
    \setlength{\tabcolsep}{4.9pt}
    \centering
    \caption{Properties of Incomplete Gamma Kernels $ K_{\Gamma}(\vec{x} \given p, \sigma^2) $ in $ \mathbb{R}^d $ for $ { p > 0 } $}
    \begin{tabular}{c R{0.425\linewidth} L{0.435\linewidth}}
        \toprule
        \multirow{4}{*}[-0.5em]{\rotatebox[origin=c]{90}{Distribution}} & Mean & $ \vec{0} $ \\
        & Covariance & $ \frac{d + p}{d + 2} \, \sigma^2 \, \mat{\identity} $ \\
        \cmidrule{2-3}
        & Characteristic function & $ {\textstyle \hyp1f1( \frac{d + p}{2}, \frac{d + 2}{2}, - \frac{\sigma^2 \norm{\vec{\omega}}^2}{2}) } $ \\
        & Moment-generating function & $ {\textstyle \hyp1f1( \frac{d + p}{2}, \frac{d + 2}{2}, \frac{\sigma^2 \norm{\vec{\omega}}^2}{2}) } $ \\
        \midrule
        \multirow{5}{*}{\rotatebox[origin=c]{90}{Mean Shift}} & Differentiable profile & \checkmark \ except for $ x = 0 $ if $ p \in (0, 2) $ \\
        & Strictly decreasing profile & \checkmark \\
        & Convex profile & \checkmark \ for $ p \in (0, 2] $ \\
        & Analytic & \checkmark \\
        & Bounded & \checkmark \\
        \midrule
        \multirow{2}{*}{\rotatebox[origin=c]{90}{Other}} & Completely monotonic profile & \checkmark \ for $ p \in (0, 2] $ \\
        & Strictly positive definite & \checkmark \ for $ p \in (0, 2] $ \\
        \bottomrule
    \end{tabular}
    \label{tab:kernel_properties}
\end{table}

\section{Kernel Properties}
\label{sec:kernel_properties}

In the following, we derive several theoretical properties of the family of incomplete gamma kernels in $ \mathbb{R}^d $ which are summarized in Table~\ref{tab:kernel_properties} and later leveraged in the applications (see Section~\ref{sec:applications}).

\subsection{Characteristic Function and Fourier Transform}
\label{sec:characteristic_function_and_fourier_transform}

Feature preservation is a highly relevant aspect in the development of denoising approaches (see Fig.~\ref{fig:gamma_kernels_comparison}) and will later be taken into account in the definition of density weighting schemes (see Section~\ref{sec:wlop_density_weights}), loss functions (see Section~\ref{sec:robust_loss_functions}), and neural network priors (see Section~\ref{sec:neural_network_priors}).
In order to gain a deeper understanding of the proposed kernel family $ K_{\Gamma} $ in this context, we are interested in its characteristic function $ \varphi_{\Gamma}(\vec{\omega}) $ which can also be interpreted as the Fourier transform $ \mathcal{F} $ of $ K_{\Gamma} $ at the angular frequency $ { \vec{\omega} \in \mathbb{R}^d } $.

First, we can apply the relation between the $ d $\mbox{-}dimensional Fourier transform of a radially symmetric function $ f(\vec{x}) $ in terms of the Hankel transform of order $ { d / 2 - 1 } $ of the function $ { \norm{\vec{x}}^{d / 2 - 1} \, f(\norm{\vec{x}}) } $~\cite{stein1971introduction} to reduce the dimensionality of the integral to the radial component
\begin{equation}
\begin{split}
    \varphi_{\Gamma}(\vec{\omega})
    & =
    \mathcal{F}\left[ K_{\Gamma}(\vec{x} \given p, \sigma^2) \right](\vec{\omega})
    =
    \int_{\mathbb{R}^d} K_{\Gamma}(\vec{x} \given p, \sigma^2) \, \euler^{\imag \dotproduct{\vec{\omega}}{\vec{x}}} \, \mathrm{d} \vec{x}
    \\
    & =
    c_{K_{\Gamma}} \frac{(2 \pi)^{\frac{d}{2}}}{\norm{\vec{\omega}}^{\frac{d}{2} - 1}} \int_{0}^{\infty} {\textstyle \Gamma(\frac{p}{2}, \frac{r^2}{2 \sigma^2}) } \, \bessel_{\frac{d}{2} - 1}(\norm{\vec{\omega}} \, r) \, r^{\frac{d}{2}} \, \mathrm{d} r
\end{split}
\end{equation}
where $ \bessel_q(x) $ denotes the \emph{Bessel function of the first kind} of order $ q $.
This integral has the closed-form solution (see the Appendix for a more detailed derivation):
\begin{equation}
\begin{split}
    & c_{K_{\Gamma}} \frac{(2 \pi)^{\frac{d}{2}}}{\norm{\vec{\omega}}^{\frac{d}{2} - 1}} \int_{0}^{\infty} {\textstyle \Gamma(\frac{p}{2}, \frac{r^2}{2 \sigma^2}) } \, \bessel_{\frac{d}{2} - 1}(\norm{\vec{\omega}} \, r) \, r^{\frac{d}{2}} \, \mathrm{d} r
    \\
    = \ &
    c_{K_{\Gamma}} (2 \pi \sigma^2)^{\frac{d}{2}} \frac{\Gamma(\frac{d + p}{2})}{\Gamma(\frac{d + 2}{2})} \, { \textstyle \hyp1f1( \frac{d + p}{2}, \frac{d + 2}{2}, - \frac{\sigma^2 \norm{\vec{\omega}}^2}{2} ) }
    \\
    = \ &
    { \textstyle \hyp1f1( \frac{d + p}{2}, \frac{d + 2}{2}, - \frac{\sigma^2 \norm{\vec{\omega}}^2}{2} ) }
\end{split}
\end{equation}
Therefore, the characteristic function of the incomplete gamma kernel can be written in terms of the \emph{confluent hypergeometric function of the first kind} $ \hyp1f1 $.
Fig.~\ref{fig:kernel_fourier} shows a comparison between the Gaussian kernel ($ { p = 2 } $) and the LOP kernel ($ { p = 1 } $) both in spatial and in frequency domain.

If we consider the special case $ { \hyp1f1(a, a, x) = \euler^x } $, we can observe that this result is consistent with the Fourier transform of the Gaussian kernel.
Furthermore as $ { d \rightarrow \infty } $, the entire family of localized $ L_p $ kernel estimators converges to the $ L_2 $ estimator since distances become increasingly similar in higher dimensions due to the curse of dimensionality.

\begin{figure}
    \centering
    \includegraphics[width=\linewidth, height=0.5\textheight, keepaspectratio]{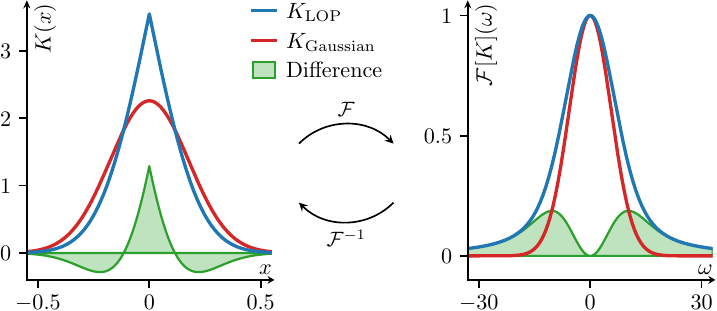}
    \caption{Comparison of LOP and Gaussian kernels in spatial and frequency domain. Filtering with the LOP kernel $ K_{\mathrm{LOP}} $ better preserves higher frequency information.}
    \label{fig:kernel_fourier}
\end{figure}

\subsection{Moment-generating Function}
\label{sec:moment_generating_function}

Another closely related and useful quantity to consider is the moment-generating function $ M_{\Gamma} $ of the kernel $ K_{\Gamma} $ which can be used to compute its mean vector $ { \vec{\mu}_{\Gamma} \in \mathbb{R}^d } $ and its covariance matrix $ { \mat{\Sigma}_{\Gamma} \in \mathbb{R}^{d \times d} } $.
Especially the insights about the magnitude of $ \mat{\Sigma}_{\Gamma} $ are crucial to define consistent density weights (see Section~\ref{sec:wlop_density_weights}) and an accurate CLOP approximation (see Section~\ref{sec:clop_kernel_approximation}).
Although there is a direct connection to the characteristic function in terms of
\begin{equation}
    M_{\Gamma}(\vec{\omega})
    =
    \varphi_{\Gamma}(- \imag \, \vec{\omega})
    =
    { \textstyle \hyp1f1( \frac{d + p}{2}, \frac{d + 2}{2}, \frac{\sigma^2 \norm{\vec{\omega}}^2}{2} ) }
\end{equation}
it does not necessarily exist in general, so we have to prove this property for all $ { \vec{\omega} \in \mathbb{R}^d } $.
For this purpose, we repeatedly apply the comparison theorem of calculus to derive finite bounds of the integral.
We first observe that $ { M_{\Gamma}(\vec{\omega}) > 0 } $ since the integrand consisting of $ K_{\Gamma} $ and an exponential term is positive.
To obtain an upper bound of $ M_{\Gamma} $, we switch to spherical coordinates and bound $ { \cos(\angle(\vec{\omega}, \vec{x})) \le 1 } $, where $ \angle(\vec{\omega}, \vec{x}) $ denotes the angle between $ \vec{\omega} $ and $ \vec{x} $, to decouple the radial component from the angular one:
\begin{equation}
\begin{split}
    M_{\Gamma}(\vec{\omega})
    & =
    \int_{\mathbb{R}^d} K_{\Gamma}(\vec{x} \given p, \sigma^2) \, \euler^{\dotproduct{\vec{\omega}}{\vec{x}}} \, \mathrm{d} \vec{x}
    \\
    & =
    c_{K_{\Gamma}} \int_{\Omega} \int_{0}^{\infty} {\textstyle \Gamma(\frac{p}{2}, \frac{r^2}{2 \sigma^2}) } \, r^{d - 1} \, \euler^{\norm{\vec{\omega}} r \cos(\angle(\vec{\omega}, \vec{x}))} \, \mathrm{d} r \, \mathrm{d} \vec{\Omega}
    \\
    & \le
    c_1 \int_{0}^{\infty} {\textstyle \Gamma(\frac{p}{2}, \frac{r^2}{2 \sigma^2}) } \, r^{d - 1} \, \euler^{\norm{\vec{\omega}} r} \, \mathrm{d} r
\end{split}
\end{equation}
For brevity, we put finite terms into constants $ c_i $.
Next, we combine the two individual upper bounds
\begin{equation}
    \Gamma(a, x)
    \le
    \begin{cases}
       a \, x^{a - 1} \, \euler^{-x}, & a \in [1, \infty), x \in [a, \infty) \text{ \cite{natalini2000inequalities}} \\
       \phantom{a} \, x^{a - 1} \, \euler^{-x}, & a \in (0, 1], x \in (0, \infty) \text{ (partial int.)}\\
    \end{cases}
\end{equation}
and apply them after splitting the integral at $ { r_0 = \max(1, p / 2) } $.
Furthermore, we simplify the expression by completing the square in the exponential term:
\begin{equation}
\begin{split}
    & c_1 \int_{0}^{\infty} {\textstyle \Gamma(\frac{p}{2}, \frac{r^2}{2 \sigma^2}) } \, r^{d - 1} \, \euler^{\norm{\vec{\omega}} r} \, \mathrm{d} r
    \\
    \le \ &
    c_1 \, c_2 + c_1 \int_{r_0}^{\infty} {\textstyle r_0 \, (\frac{r^2}{2 \sigma^2})^{\frac{p}{2} - 1} \, \euler^{- \frac{r^2}{2 \sigma^2}} } \, r^{d - 1} \, \euler^{\norm{\vec{\omega}} r} \, \mathrm{d} r
    \\
    = \ &
    c_1 \, c_2 + c_1 \, c_3 \int_{r_0}^{\infty} r^{p + d - 3} \, \euler^{- \frac{(r - \sigma^2 \norm{\vec{\omega}})^2}{2 \sigma^2}} \, \mathrm{d} r
\end{split}
\end{equation}
Since $ { r \in [1, \infty) } $, the remaining polynomial can be bound by a higher order $ { k = \max(0, \lceil p + d - 3 \rceil) \in \mathbb{N}_0 } $.
Therefore, this integral describes the (incomplete) $ k $-th raw moment of a 1D normal distribution and is finite for any $ { k \in \mathbb{N}_0 } $ which, in turn, proves that $ { M_{\Gamma}(\vec{\omega}) < \infty } $ exists.

\subsubsection{Mean}
We can now use the moment-generating function $ M_{\Gamma} $ to directly compute all raw moments of the kernel $ K_{\Gamma} $ by evaluating the respective derivative at $ { \vec{\omega} = \vec{0} } $.
For the mean vector $ { \vec{\mu}_{\Gamma} \in \mathbb{R}^d } $ of $ K_{\Gamma} $, we consider the first-order derivative
\begin{equation}
    \frac{\partial}{\partial \vec{\omega}} M_{\Gamma}(\vec{\omega})
    =
    { \textstyle \frac{d + p}{d + 2} \, \sigma^2 \, \hyp1f1( \frac{d + p + 2}{2}, \frac{d + 4}{2}, \frac{\sigma^2 \norm{\vec{\omega}}^2}{2}) } \, \vec{\omega}
\end{equation}
and get $ { \vec{\mu}_{\Gamma} = \frac{\partial}{\partial \vec{\omega}} M_{\Gamma}(\vec{0}) = \vec{0} } $ as the expected result for a radially symmetric kernel.

\subsubsection{Covariance}
Similarly, we compute the second-order derivative
\begin{equation}
\begin{split}
    \frac{\partial^2}{\partial \vec{\omega} \, \partial \vec{\omega}^{\transpose}} M_{\Gamma}(\vec{\omega})
    & =
    { \textstyle \frac{d + p}{d + 2} \, \sigma^2 \, \left[ \hyp1f1( \frac{d + p + 2}{2}, \frac{d + 4}{2}, \frac{\sigma^2 \norm{\vec{\omega}}^2}{2}) \, \mat{\identity} \right. }
    \\
    & +
    { \textstyle \left. \frac{d + p + 2}{d + 4} \, \sigma^2 \, \hyp1f1( \frac{d + p + 4}{2}, \frac{d + 6}{2}, \frac{\sigma^2 \norm{\vec{\omega}}^2}{2}) \, \vec{\omega} \vec{\omega}^{\transpose} \right] }
\end{split}
\end{equation}
and obtain the covariance matrix $ { \mat{\Sigma}_{\Gamma} \in \mathbb{R}^{d \times d} } $ of the kernel $ K_{\Gamma} $ using the first-order and second-order raw moments as $ { \mat{\Sigma}_{\Gamma} = \frac{\partial^2}{\partial \vec{\omega} \, \partial \vec{\omega}^{\transpose}} M_{\Gamma}(\vec{0}) - \vec{\mu}_{\Gamma} \vec{\mu}_{\Gamma}^{\transpose} = [(d + p) / (d + 2)] \, \sigma^2 \, \mat{\identity} } $.

\subsection{Mean Shift Properties}
\label{sec:mean_shift_properties}

\begin{figure*}
    \centering
    \includegraphics[width=\linewidth, height=0.2\textheight, keepaspectratio]{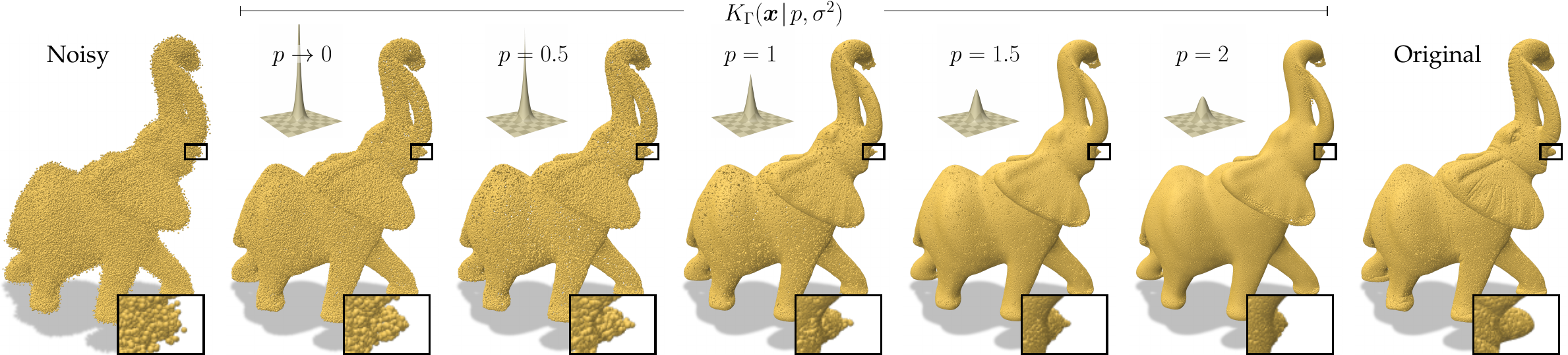}
    \caption{Exemplary point cloud denoising of the \emph{Elephant} model (\num{302458} points) with \num{30} WLOP~\cite{huang2009consolidation} iterations ($ { h = 6 } $, $ { \mu = 0.4 } $) for different incomplete gamma kernels $ K_{\Gamma} $ using varying $ { p \in (0, 2] } $ and fixed $ { \sigma^2 = 1 / 32 } $. The model has been corrupted with $ { \sigma_{\mathrm{noise}} = 0.3 } $ ($ \SI{80}{\percent} $ points) and $ { \sigma_{\mathrm{outlier}} = 1.5 } $ Gaussian noise ($ \SI{20}{\percent} $ points) respectively to account for both typical sensor noise and heavy outliers. Higher $ p $\mbox{-}norms result in more regular but oversmoothed point distributions whereas lower values better preserve features. Unit of $ h, \sigma_{\mathrm{noise}}, \sigma_{\mathrm{outlier}} $: $ [\% \text{ BB diagonal}] $.}
    \label{fig:gamma_kernels_comparison}
\end{figure*}

In addition to the distribution-specific properties above, we can get further insights into the kernel family by exploiting the comprehensive theory that has been developed for the Mean Shift algorithm~\cite{yamasaki2019properties}.
This requires proving several additional properties including that the kernel $ K_{\Gamma} $ is bounded and analytic and that its profile $ k_{\Gamma} $ is differentiable, strictly decreasing, and convex.

\subsubsection{Differentiability, Monotonicity and Convexity}
In order to show that the profile is strictly decreasing, we consider its first-order derivative
\begin{equation}
    \frac{\mathrm{d}}{\mathrm{d} x} k_{\Gamma}(x \given p, \sigma^2)
    =
    \frac{\mathrm{d}}{\mathrm{d} x} {\textstyle \Gamma(\frac{p}{2}, \frac{x}{2 \sigma^2}) }
    =
    { \textstyle - (\frac{1}{2 \sigma^2})^{\frac{p}{2}} \, x^{\frac{p}{2} - 1} \, \euler^{- \frac{x}{2 \sigma^2}} }
\end{equation}
which is defined for all $ { x \in (0, \infty) } $ as well as for $ { x = 0 } $ if $ { p \in [2, \infty) } $.
Since the involved polynomial and exponential terms are always positive, it follows that the derivative must be negative, that is $ { \frac{\mathrm{d}}{\mathrm{d} x} k_{\Gamma}(x \given p, \sigma^2) < 0 } $, and the profile strictly decreasing.

Similarly, we see that the second-order derivative is given by
\begin{equation}
    \frac{\mathrm{d}^2}{\mathrm{d} x^2} k_{\Gamma}(x \given p, \sigma^2)
    =
    \frac{\mathrm{d}}{\mathrm{d} x} { \textstyle k_{\Gamma}(x \given p, \sigma^2) } \, \left[ \frac{\frac{p}{2} - 1}{x} - \frac{1}{2 \sigma^2} \right]
\end{equation}
where, in order to ensure that $ { \frac{\mathrm{d}^2}{\mathrm{d} x^2} k_{\Gamma}(x \given p, \sigma^2) > 0 } $, the latter term must be non-positive which is equivalent to the condition $ { x \ge (p - 2) \, \sigma^2 } $.
Since this should hold for all $ { x \in (0, \infty) } $, convexity is only guaranteed for kernels with $ { p \in (0, 2] } $ which, in particular, includes the Gaussian kernel ($ { p = 2 } $) as well as the LOP kernel ($ { p = 1 } $).

\subsubsection{Boundedness and Analyticity}
Instead of showing the properties of boundedness and analyticity for the kernel $ K_{\Gamma} $ itself, it is sufficient to show them for its profile $ k_{\Gamma} $.
Since $ k_{\Gamma} $ is non-negative and monotonically decreasing, we only have to consider the case $ { x = 0 } $.
For this value, $ \Gamma(a, x) $ reduces to the gamma function $ \Gamma(a) $ which is finite for $ { a > 0 } $.
Furthermore, analyticity directly follows from the fact that $ \Gamma(a, x) $ is holomorphic in $ { x \in (0, \infty) } $ for any fixed $ { a > 0 } $.

\subsubsection{Consequences for the LOP operator}
The aforementioned properties have several direct implications~\cite{yamasaki2019properties} on the behavior of the LOP operator (with zero repulsion) as well as to Mean Shift applied with the incomplete gamma kernel $ K_{\Gamma} $ for $ { p \in (0, 2] } $.
With the exception of the finite set of target points $ \set{P} $ where singularities are introduced in the kernel $ G_{\Gamma} $, the following properties hold:

\textbf{Non-zero Gradient.}
The gradient of the kernel density estimate $ \nabla \hat{f}_{\set{P},K_{\Gamma}} $ in \eqref{eq:kernel_density_estimate_gradient} is non-zero outside the convex hull of the target point set $ \set{P} $.
This implies that all solutions must lie within the convex hull.

\textbf{Plateau-free Density.}
In addition to non-zero gradients, the kernel density estimate on the set $ \mathbb{R}^d \tightsetminus \set{P} $ has no plateaus.
Since the set of target points $ \set{P} $ is finite, we can extend this property to the full space $ \mathbb{R}^d $.

\textbf{Non-decreasing Density Estimate.}
Another interesting subset to consider is the improvement ball $ \set{I}(\vec{q}_j^{(t)}) $ which denotes a $ d $\mbox{-}dimensional sphere centered at the point $ { \vec{q}_j^{(t)} + \vec{m}_{\set{P},G_{\Gamma}}(\vec{q}_j^{(t)}) } $ with radius $ { \smallnorm{\vec{m}_{\set{P},G_{\Gamma}}(\vec{q}_j^{(t)})} } $.
In case of the LOP operator, it follows that all points $ \vec{x} $ within the improvement ball have non-decreasing kernel density estimates $ { \hat{f}_{\set{P},K_{\Gamma}}(\vec{x}) \ge \hat{f}_{\set{P},K_{\Gamma}}(\vec{q}_j^{(t)}) } $.

\textbf{Convergence of Density Estimate Sequence.}
As a consequence of the above property, the sequence of kernel density estimates $ { \{ \hat{f}_{\set{P},K_{\Gamma}}(\vec{q}^{(t)}) \} } $ obtained via the fixed-point iteration $ { \vec{q}^{(t + 1)} = \vec{q}^{(t)} + \vec{m}_{\set{P},G_{\Gamma}}(\vec{q}^{(t)}) } $ is non-decreasing.
Furthermore, this sequence always converges.

\textbf{Convergence of Mode Estimate Sequence.}
Finally, we can conclude that the mode estimate sequence $ { \{ \vec{q}^{(t)} \} } $ converges to a single point.
Depending on the window size $ h $ and the distribution of the target points $ \set{P} $, this solution could be either a point $ { \vec{p} \in \set{P} } $ due to the singularity (for very small window sizes) or a different point $ { \vec{p} \in \mathbb{R}^d \tightsetminus \set{P} } $ in the corresponding convex hull (for larger window sizes).
While a proof for the convergence of a modified version of LOP in $ \mathbb{R}^d $ has only been shown very recently~\cite{faigenbaum2023manifold}, this provides an alternative way through the comprehensive Mean Shift theory which, though only considering the attraction term, extends and generalizes to all kernels $ K_{\Gamma} $ with $ { p \in (0, 2] } $.

\subsection{Further Properties}
\label{sec:further_properties}

Besides the theoretical results concerning basic distribution-related aspects as well as insights in the projection behavior of LOP from the perspective of Mean Shift, we want to derive further properties and, in particular, strict positive definiteness.
This opens up a broader set of applications beyond improved density weights (see Section~\ref{sec:wlop_density_weights}) and may also be relevant, \eg, in the field of Gaussian Process Regression, which, however is out of the scope of this work.

\subsubsection{Complete Monotonicity}
For this purpose, we show that the kernel profile $ k_{\Gamma} $ is completely monotonic, that is $ { (-1)^n \, \frac{\mathrm{d}^n}{\mathrm{d} x^n} k_{\Gamma}(x \given p, \sigma^2) \ge 0 } $ for all $ { n \in \mathbb{N}_0 } $ and $ { x \in (0, \infty) } $.
From the derivation of the Mean Shift properties, we already know that $ { k_{\Gamma}(x \given p, \sigma^2) > 0 } $ and $ { \frac{\mathrm{d}}{\mathrm{d} x} k_{\Gamma}(x \given p, \sigma^2) < 0 } $ holds for all $ { x \in (0, \infty) } $.
Since $ x^a $ with $ { a \le 0 } $ as well as $ \euler^{- x} $ are both completely monotonic and the product of two completely monotonic functions retains that property~\cite{schilling2012bernstein}, we can conclude that this also holds for $ k_{\Gamma}(x \given p, \sigma^2) $ with $ { p \in (0, 2] } $.

\subsubsection{Strict Positive Definiteness}
A direct consequence of the complete monotonicity of its profile is that the kernel $ K_{\Gamma} $ is a positive definite function.
Furthermore, since the profile is also not a constant function, $ K_{\Gamma} $ must be even strictly postive definite~\cite{schoenberg1938metric}.
Therefore, for any set of distinct points $ \set{P} $, the matrix
\begin{equation}
    \mat{C}
    =
    \left( K_{\Gamma}(\vec{p}_i - \vec{p}_j \given p, \sigma^2) \right)_{ij}
    \in \mathbb{R}^{\abs{\set{P}} \times \abs{\set{P}}}
    \label{eq:equation_system}
\end{equation}
is symmetric and positive definite for $ { p \in (0, 2] } $, so it can be interpreted as a Gram matrix and any linear system with respect to $ \mat{C} $ has a unique solution which can be computed by, \eg, conjugate gradient solvers.
This also directly extends to any truncated version of $ K_{\Gamma} $ where the matrix $ \mat{C} $ becomes sparse and more efficient to solve as vanishing derivatives of the truncated profile do not affect complete monotonicity.

\section{Applications}
\label{sec:applications}

Besides the application of other localized $ L_p $ estimators for point cloud denoising via the incomplete gamma kernels $ K_{\Gamma} $, as shown in Fig.~\ref{fig:gamma_kernels_comparison}, we illustrate several further applications to demonstrate the benefits of the theoretical results derived for the kernel family.

\subsection{WLOP Density Weights}
\label{sec:wlop_density_weights}

In addition to a high robustness to noise, another crucial requirement in point cloud denoising is a uniform distribution of the denoised points.
Although the introduction of an additional repulsion energy \eqref{eq:repulsion_energy} mitigates the clustering effect of the attraction term \eqref{eq:attraction_energy}, the projection is still highly dependent on the distribution of the target points $ \set{P} $.
This has been addressed in WLOP~\cite{huang2009consolidation} by computing weights $ v_i $ for each target point $ \vec{p}_i $ and $ v_j^{(t)} $ for each projection point $ \vec{q}_j^{(t)} $ as an additional regularization based on the reciprocal and ordinary density value respectively with the (unnormalized) localization kernel $ \theta $.
However, our derived theoretical properties reveal two major limitations of this particular choice:
1) Although the Gaussian localization kernel $ \theta $ could be considered a reasonable approximation of the actual kernel $ K_{\mathrm{LOP}} $ given in \eqref{eq:lop_kernel}, it significantly differs in terms of the frequency spectrum (see Section~\ref{sec:characteristic_function_and_fourier_transform} and Fig.~\ref{fig:kernel_fourier}) as well as covariance (see Section~\ref{sec:moment_generating_function}) and leads to oversmoothing of high-frequency density information and, hence, a lacking preservation of fine-scale details;
2) taking the reciprocal to invert the density of $ \vec{p}_i $ ignores the dependencies between the weights which corresponds to the assumption of constant density in a window of size $ h $.
In order to achieve a more accurate normalization, we propose two novel weighting schemes.

\textbf{Simple Scheme.}
A simple extension to the WLOP weights keeps the assumption of the latter limitation and addresses only the former one by applying the actual kernel $ K_{\mathrm{LOP}} $, that is estimating the weights
\begin{equation}
    v_i
    =
    \frac{1}{\hat{f}_{\set{P}, K_{\mathrm{LOP}}}(\vec{p}_i)}
    , \quad
    v_j^{(t)}
    =
    \hat{f}_{\set{Q}^{(t)}, K_{\mathrm{LOP}}}(\vec{q}_j^{(t)})
    \label{eq:simple_scheme}
\end{equation}
via the kernel density estimate \eqref{eq:kernel_density_estimate} of the point clouds $ \set{P} $ and $ \set{Q}^{(t)} $ respectively.
This scheme can be easily integrated into existing applications of WLOP as it only involves a different kernel function in the overall weight computation.

\textbf{Full Scheme.}
To address both limitations, we consider the kernel density estimate $ \hat{f}_{\set{P}, K_{\mathrm{LOP}}} $ with weights $ v_i $ applied to each term.
We want to enforce constant density at the points $ \vec{p}_i $ which can be formulated as a linear optimization problem in matrix form:
\begin{equation}
    \frac{1}{\abs{\set{P}} \, h^d} \, \left( {\textstyle K_{\mathrm{LOP}}( \frac{\vec{p}_i - \vec{p}_j}{h}) } \right)_{ij} \, (\dots, v_i, \dots)^{\transpose} = \vec{1}
    \label{eq:full_scheme}
\end{equation}
This corresponds to radial basis function (RBF) interpolation with a Gram matrix similar to \eqref{eq:equation_system} and we can obtain a unique solution since $ K_{\mathrm{LOP}} $ is strictly positive definite (see Section~\ref{sec:further_properties}).
Furthermore, we truncate the kernel at $ { h / 2 } $ to drastically reduce the memory requirements of the matrix and use a sparse conjugate gradient solver.
In case of the projection points $ \vec{q}_j^{(t)} $, we consider the inverse of the matrix which leads to the same weights as in the simple scheme.

\subsection{CLOP Kernel Approximation}
\label{sec:clop_kernel_approximation}

\begin{table}
    \renewcommand{\arraystretch}{1.1}
    \centering
    \caption{Parameter Sets of Kernel Approximation $ \hat{K}_{\mathrm{LOP}} $}
    \begin{tabular}{
        S[table-number-alignment=center, table-format=1.0]
        S[table-number-alignment=center, table-format=2.3]
        S[table-number-alignment=center, table-format=1.5, group-minimum-digits=6]
        S[table-number-alignment=center, table-format=2.3]
        S[table-number-alignment=center, table-format=1.5, group-minimum-digits=6]
        S[table-number-alignment=center, table-format=2.3]
        S[table-number-alignment=center, table-format=1.5, group-minimum-digits=6]
    }
        \toprule
        & \multicolumn{2}{c}{CLOP~\cite{preiner2014continuous}} & \multicolumn{2}{c}{Ours} & \multicolumn{2}{c}{Ours (Consistent)} \\
        \cmidrule(lr){2-3} \cmidrule(lr){4-5} \cmidrule(lr){6-7}
        {$ k $} & {$ \hat{w}_k $} & {$ \hat{\sigma}_k $} & {$ \hat{w}_k $} & {$ \hat{\sigma}_k $} & {$ \hat{w}_k $} & {$ \hat{\sigma}_k $} \\
        \midrule
        1 & 97.761 & 0.01010 & 61.509 & 0.02102 & 46.409 & 0.03118 \\
        2 & 29.886 & 0.03287 & 11.932 & 0.07289 & 9.635 & 0.10582 \\
        3 & 11.453 & 0.11772 & 5.069 & 0.15700 & 2.674 & $ \sqrt{1 / 32} $ \\
        \bottomrule
    \end{tabular}
    \label{tab:kernel_approximation}
\end{table}

While the major aspect of ensuring a uniform distribution of the projected point set can be addressed with an appropriate weighting as shown in Section~\ref{sec:wlop_density_weights}, obtaining these results efficiently becomes challenging especially on large datasets as all target points $ \set{P} $ will be taken into account for denoising.
For this purpose, CLOP~\cite{preiner2014continuous} first computes a more compact representation of the points $ \set{P} $ in terms of a smaller set of normal distributions $ { \set{P}_{\set{N}} = \{ (w_i, \vec{\mu}_i, \mat{\Sigma}_i) \} } $ with weights $ { w_i \in \mathbb{R} } $, mean vectors $ { \vec{\mu}_i \in \mathbb{R}^3 } $, and local covariance matrices $ { \mat{\Sigma}_i \in \mathbb{R}^{3 \times 3} } $ and then extends the discrete attraction energy to the continuous space.
However, since the integral in the respective update step cannot be directly solved, the kernel $ { \alpha(\norm{\vec{x}}) } $ is approximated by a radially symmetric Gaussian mixture model $ { \hat{\alpha}(\vec{x}) = (1 / h) \sum_{k = 1}^{3} \hat{w}_k \, \hat{c}_k \, \mathcal{N}(\vec{x} / h \given \vec{0}, \hat{\sigma}_k^2  \, \mat{\identity}) } $ consisting of three components with fitted parameters $ { \{ (\hat{w}_k, \hat{\sigma}_k) \in \mathbb{R} \times \mathbb{R} \} } $ and dimension-dependent constants $ { \hat{c}_k = \smallabs{2 \pi \hat{\sigma}_k^2 \, \mat{\identity}}^{1 / 2} } $.
In the context of Mean Shift, this implies that the kernel $ G_{\mathrm{LOP}} $ is in fact approximated which directly allows us to derive
\begin{equation}
    \hat{K}_{\mathrm{LOP}}(\vec{x})
    =
    \frac{\sum_{k = 1}^{3} \hat{\sigma}_k^2 \, \hat{w}_k \, \hat{c}_k \, \mathcal{N}(\vec{x} \given \vec{0}, \hat{\sigma}_k^2 \, \mat{\identity})}{\sum_{k = 1}^{3} \hat{\sigma}_k^2 \, \hat{w}_k \, \hat{c}_k \, \phantom{\mathcal{N}(\vec{x} \given \vec{0}, \hat{\sigma}_k^2 \, \mat{\identity})}}
    \label{eq:lop_kernel_approximation}
\end{equation}
as an approximation of the kernel $ K_{\mathrm{LOP}} $ in $ \mathbb{R}^d $ with the same set of fitted parameters $ { \{ (\hat{w}_k, \hat{\sigma}_k) \} } $.

\textbf{Kernel Fit.}
Finding the optimal parameter set is highly challenging due to the singularity of $ G_{\mathrm{LOP}} $ at $ { x = 0 } $ and, thereby, the unbounded ratio between the smallest and largest sampling value in the half-open fitting interval $ { (0, 1] } $ for $ { h = 1 } $.
In contrast, we directly optimize on the kernel $ K_{\mathrm{LOP}} $ which does not suffer from these limitations.
We fix the parameter $ { w_3 = 1 } $ to constrain the remaining degree of freedom and obtain the solution from $ 10^7 $ uniformly sampled points in the interval $ { [0, 1] } $ via the Levenberg-Marquardt algorithm (see Table~\ref{tab:kernel_approximation}).
Although the LOP operator is scale-invariant in terms of the kernel $ \alpha $, we nevertheless estimate a global scaling factor for the weights $ \hat{w}_k $ via Levenberg-Marquardt optimization in the interval $ { (0.01, 1] } $ for a better comparability with CLOP.

\textbf{Consistent Fit.}
In addition to the derivation of the kernel approximation $ \hat{K}_{\mathrm{LOP}} $, we can futher apply the insights about the covariance of $ K_{\mathrm{LOP}} $ (see Section~\ref{sec:moment_generating_function}).
From \eqref{eq:lop_kernel_approximation}, we can also see that the covariance matrix $ { \hat{\mat{\Sigma}}_{\mathrm{LOP}} \in \mathbb{R}^{d \times d} } $ of the approximation consists of a convex combination of the parameters $ \hat{\sigma}_k^2 $:
\begin{equation}
    \hat{\mat{\Sigma}}_{\mathrm{LOP}}
    =
    \frac{\sum_{k = 1}^{3} \hat{w}_k \, \hat{\sigma}_k^{d + 4}}{\sum_{k = 1}^{3} \hat{w}_k \, \hat{\sigma}_k^{d + 2}} \, \mat{\identity}
    \label{eq:lop_kernel_approximation_variance}
\end{equation}
In the limit $ { d \to \infty } $, this combination degenerates to $ { \hat{\mat{\Sigma}}_{\mathrm{LOP}} \to \max_k{\hat{\sigma}_k^2} \, \mat{\identity} } $ which is similar to the maximum norm $ L_{\infty} $ being the limit of the $ L_p $ norms.
Therefore, we can enforce an additional consistency constraint in the parameter optimization process by fixing the parameter $ { \hat{\sigma}_3 = \sqrt{1 / 32} } $ such that $ \hat{\mat{\Sigma}}_{\mathrm{LOP}} $ matches the expected covariance $ { \mat{\Sigma}_{\mathrm{LOP}} = 1 / 32 \, [(d + 1) / (d + 2)] \, \mat{\identity} \to 1 / 32 \, \mat{\identity}} $.

\subsection{Robust Loss Functions}
\label{sec:robust_loss_functions}

\begin{figure}[b]
    \centering
    \includegraphics[width=\linewidth, height=0.5\textheight, keepaspectratio]{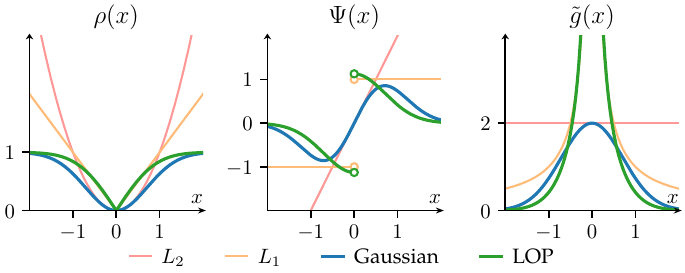}
    \caption{Comparison of LOP and Gaussian M\mbox{-}estimators for $ { \sigma^2 = 1 / 2 } $. Due to the close relation to Mean Shift, these robust loss functions $ \rho $ do not only share the shape of the corresponding kernels $ K $ but also have similar properties and form localized versions of the common global $ L_2 $ and $ L_1 $ loss functions.}
    \label{fig:gamma_loss}
\end{figure}

\begin{figure*}
    \centering
    \includegraphics[width=\linewidth, height=0.5\textheight, keepaspectratio]{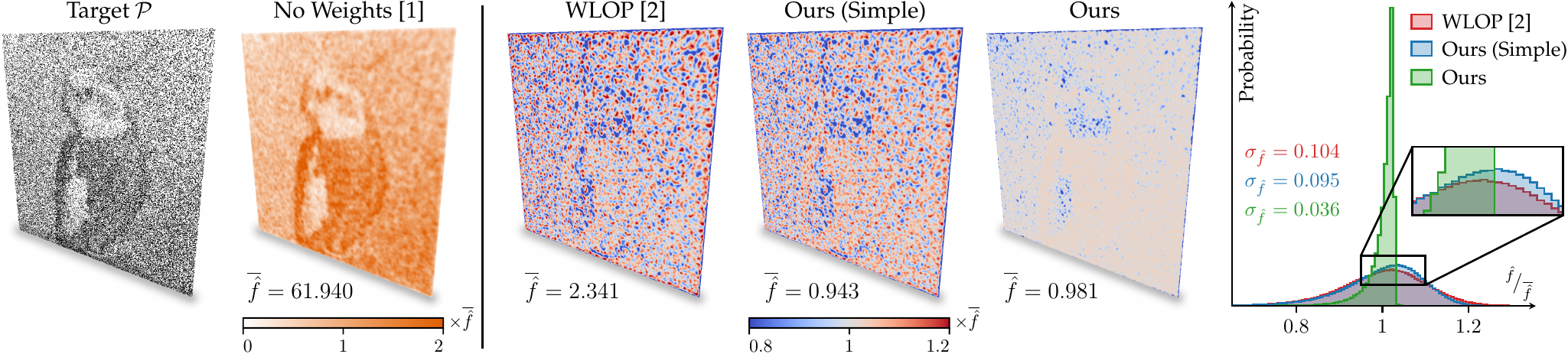}
    \caption{Kernel density estimate $ \hat{f} $ for $ { h = 3 } $ of a planar target surface patch $ \set{P} $ (\num{74000} points) that is sampled inversely proportional to the intensity of the \emph{Bird} image. The regularity $ \sigma_{\set{Q}} $ of any projected point set $ \set{Q} $ onto this target directly depends on the uniformity of $ \hat{f} $. Whereas WLOP~\cite{huang2009consolidation} and our simple weighting scheme cannot fully remove high-frequency variations, our full weighting scheme leads to a significantly better normalization and more uniform density. Unit of $ h $: $ [\% \text{ BB diagonal}] $.}
    \label{fig:weights_density}
\end{figure*}

\begin{figure}
    \centering
    \includegraphics[width=\linewidth, height=0.5\textheight, keepaspectratio]{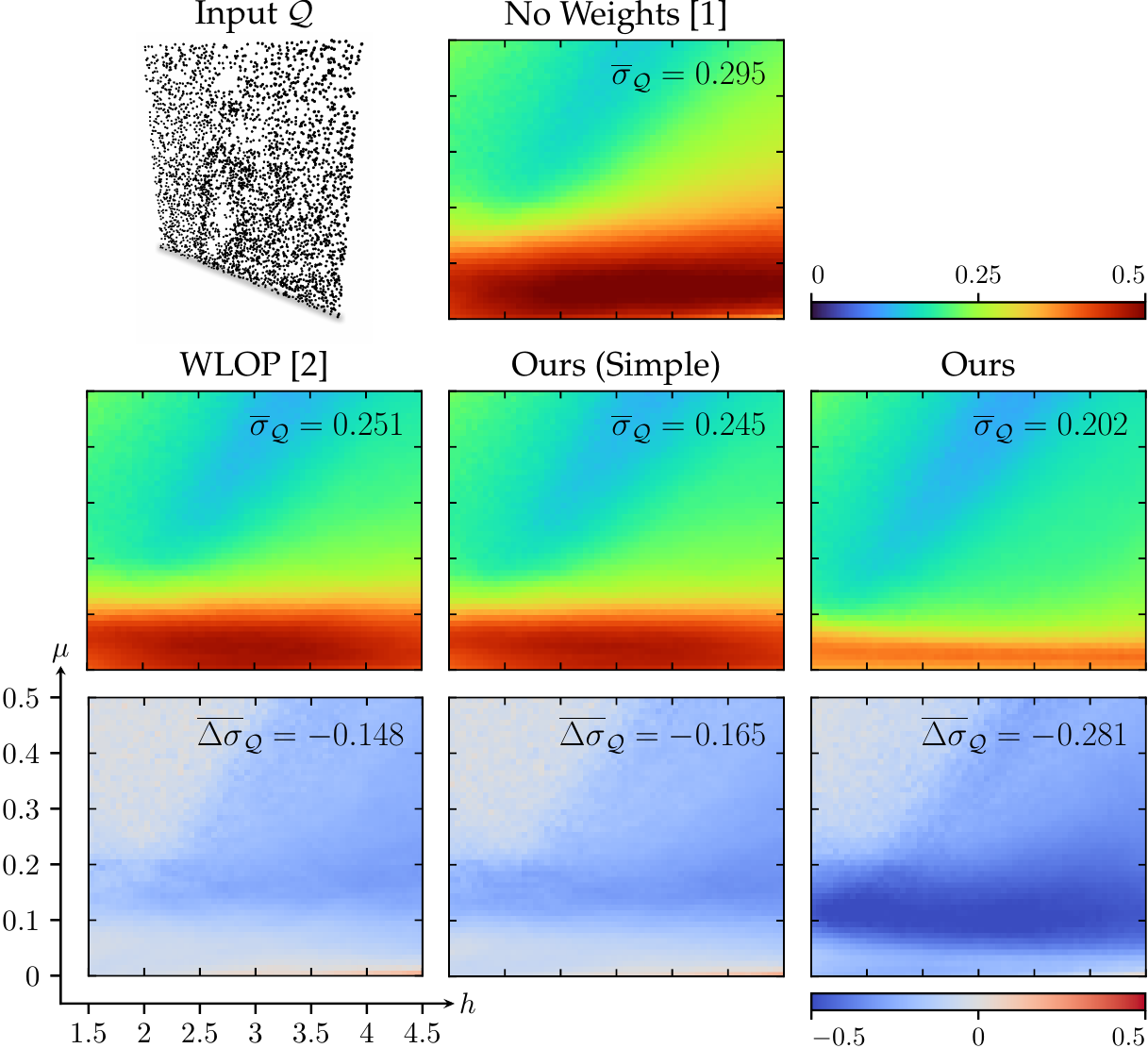}
    \caption{Regularity $ \sigma_{\set{Q}} $ of the input subset $ \set{Q} $ (\num{3700} points) projected on a planar target surface patch $ \set{P} $ (\num{74000} points) that is sampled inversely proportional to the intensity of the \emph{Bird} image. Due to the better density normalization, our weighting schemes further improve the regularity across various combinations of the window size $ h $ and the repulsion weight $ \mu $. Unit of $ h $ and $ \sigma_{\set{Q}} $: $ [\% \text{ BB diagonal}] $.}
    \label{fig:weights_regularity}
\end{figure}

A closely related, yet slightly different task is mesh denoising where the quality of the noisy vertex set (similar to point sets in point cloud denoising) should be improved while preserving the mesh connectivity and avoiding self-intersections of the faces.
Respective approaches tackle this problem in a two-stage approach where the face normals are initially denoised and subsequently used as a guidance in the second stage to consistently adjust the vertex positions.
Obtaining a reliable estimate of a denoised normal $ { \vec{n} \in \mathbb{R}^3 } $ can be performed by leveraging gradient descent or M\mbox{-}estimation from the broad field of robust statistics~\cite{hampel1986robust} which is also related to the concept of anisotropic diffusion~\cite{black1998robust}.
Here, the objective function
\begin{equation}
    L(\vec{n})
    =
    \sum_i \rho(\norm{\vec{n}_i - \vec{n}})
\end{equation}
defined with a robust loss function $ \rho $ is considered and a solution can be found based on the corresponding influence function $ { \Psi(x) = \frac{\mathrm{d}}{\mathrm{d} x} \rho(x) } $ and anisotropic weight function $ { \tilde{g}(x) = \Psi(x) / x } $~\cite{yadav2021surface}:
\begin{equation}
    \vec{n}^{(t + 1)}
    =
    \frac{\sum_i \tilde{g}(\smallnorm{\vec{n}_i - \vec{n}^{(t)}}) \, \vec{n}^{(t)}}{\norm{\sum_i \tilde{g}(\smallnorm{\vec{n}_i - \vec{n}^{(t)}}) \, \vec{n}^{(t)}}}
    \label{eq:normal_update_step}
\end{equation}
This result is closely related to the derivation of Mean Shift and shares many properties with it~\cite{comaniciu2002mean}.
While the Gaussian loss is a well-known choice in this context, we can generalize it to the family of \emph{incomplete gamma losses} along with the respective influence and anisotropic weight functions and corresponding frequency-related properties (see Section~\ref{sec:characteristic_function_and_fourier_transform}):
\begin{align}
    \rho_{\Gamma}(x \given p, \sigma^2)
    & =
    \frac{1}{\Gamma(\frac{p}{2})} \, {\textstyle \gamma(\frac{p}{2}, \frac{x^2}{2 \sigma^2}) }
    \\
    \Psi_{\Gamma}(x \given p, \sigma^2)
    & =
    \frac{2}{(2 \sigma^2)^{\frac{p}{2}} \, \Gamma(\frac{p}{2})} \, \abs{x}^{p - 2} \, \euler^{- \frac{x^2}{2 \sigma^2}} \, x
    \\
    \tilde{g}_{\Gamma}(x \given p, \sigma^2)
    & =
    \frac{2}{(2 \sigma^2)^{\frac{p}{2}} \, \Gamma(\frac{p}{2})} \, \abs{x}^{p - 2} \, \euler^{- \frac{x^2}{2 \sigma^2}}
\end{align}
Here, the losses $ \rho_{\Gamma} $ are built upon the \emph{lower incomplete gamma function} $ { \gamma(a, x) = \int_{0}^{x} t^{a - 1} \, \euler^{-t} \, \mathrm{d} t } $ which is connected to the upper incomplete gamma function via the relation $ { \gamma(a, x) + \Gamma(a, x) = \Gamma(a) } $.
By choosing $ { p = 1 } $ and applying the identity $ { \gamma(1 / 2, x) = \sqrt{\pi} \, \erf(\sqrt{x}) } $, we get the LOP loss
\begin{equation}
    \rho_{\mathrm{LOP}}(x \given \sigma^2)
    =
    {\textstyle \erf(\frac{\abs{x}}{\sqrt{2 \sigma^2}}) }
    \label{eq:lop_loss}
\end{equation}
where $ \erf $ denotes the \emph{error function} and is related to its complementary counterpart via $ { \erfc(x) = 1 - \erf(x) } $.
Considering $ { \sigma^2 = 1 / 32 } $, the relation $ { \tilde{g}_{\mathrm{LOP}}(x \given 1 / 32) \propto g_{\mathrm{LOP}}(x^2) } $ further highlights the close similarity to Mean Shift.
Due to the broad applicability of the concepts of robust statistics~\cite{hampel1986robust}, \ie the definition and optimization of loss functions, to traditional approaches, machine learning, and many other fields such as Mean Shift, our losses $ \rho_{\Gamma} $ may also be highly relevant for deep learning techniques and extend previous generalizations~\cite{barron2019general}.
Fig.~\ref{fig:gamma_loss} shows a comparison between the Gaussian M\mbox{-}estimator ($ { p = 2 } $) and the LOP M\mbox{-}estimator ($ { p = 1 } $).

\subsection{Neural Network Priors}
\label{sec:neural_network_priors}

In the context of learning-based point cloud denoising, many approaches rely on clean groundtruth data in order to learn how various types of synthetically generated noise such as Gaussian, uniform, or simulated sensor noise can be effectively removed from the given point set.
However, they may not necessarily generalize to real-world data exhibiting different possibly systematic yet unknown noise patterns and, hence, lead to less reliable and accurate results.
For this purpose, TotalDenoising~\cite{hermosilla2019total} formulates the learning objective in an unsupervised manner, which only requires noisy input data without further knowledge about the noise itself or corresponding clean data.
Given noisy points $ { \vec{p} \in \mathbb{R}^3 } $, the parameters $ \vec{\Theta} $ of a neural network $ f_{\vec{\Theta}} $ are optimized such that the observations $ \vec{p} $ are projected to the unknown surface $ \set{S} $ represented by the distribution of modes $ p(\vec{x} \given \set{S}) $.
In particular, this can be formulated by the objective function
\begin{equation}
    L(\vec{\Theta}) = \mathbb{E}_{\vec{p} \sim p(\vec{x} \given \set{S})} \, \mathbb{E}_{\vec{q} \sim q(\vec{x} \given \vec{p})} \, \rho(f_{\vec{\Theta}}(\vec{p}, \vec{q}))
\end{equation}
where $ \rho $ denotes a loss function, \eg the commonly used $ L_2 $ loss or other choices such as robust losses (see Section~\ref{sec:robust_loss_functions}), and $ \vec{q} $ denotes the generated samples during training.
Since any point $ { \vec{x} \in \set{S} } $ on the surface is a valid target onto which the noisy input $ \vec{p} $ can be projected, the unsupervised training procedure will not converge to a unique solution.
Thus, the prior distribution
\begin{equation}
    q(\vec{x} \given \vec{p}) = p(\vec{x} \given \set{S}) \, K(\mat{W}(\vec{x} - \vec{p}))
    \label{eq:network_prior}
\end{equation}
regularizes this problem by convolving the distribution $ p(\vec{x} \given \set{S}) $ with a smoothing kernel $ K $ and a diagonal weight matrix $ { \mat{W} \in \mathbb{R}^{3 \times 3} } $ to favor the closest mode as the unique solution for projection.
This convolution-based regularization is similar to the formulation of the kernel density estimate \eqref{eq:kernel_density_estimate} in the Mean Shift approach.
Therefore, instead of using a Gaussian kernel as in the original TotalDensoising approach, we can apply the gained insights of our kernel family $ K_{\Gamma} $ regarding feature preservation (see Section~\ref{sec:characteristic_function_and_fourier_transform} and Fig.~\ref{fig:kernel_fourier}) and use the LOP kernel $ K_{\mathrm{LOP}} $ given in \eqref{eq:lop_kernel} in the prior distribution.

%-------------------------------------------------------------------------
\section{Experimental Results}

In the following, we demonstrate the effectiveness of our proposed extensions in Section~\ref{sec:applications} that are derived from the theoretical properties of the kernel family (see Section~\ref{sec:kernel_properties}).

\subsection{Evaluation of WLOP Density Weights}
\label{sec:evaluation_of_wlop_density_weights}

\begin{figure}
    \centering
    \includegraphics[width=\linewidth, height=0.5\textheight, keepaspectratio]{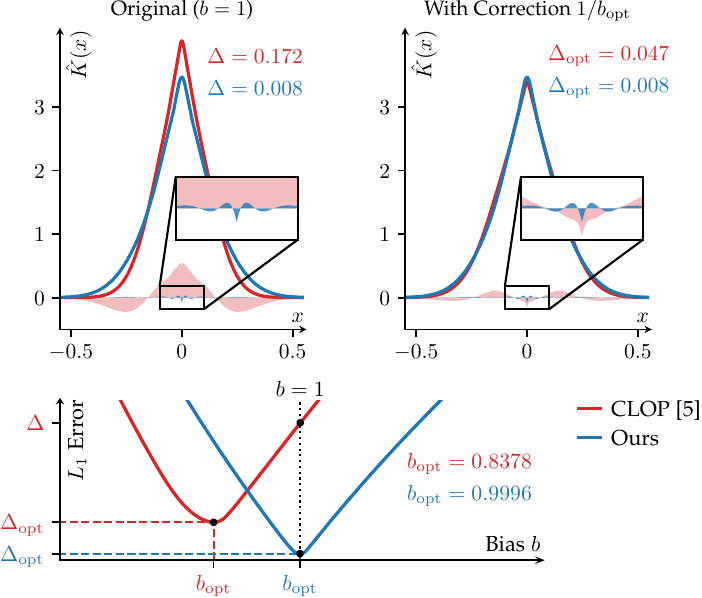}
    \caption{Analysis of bias in the width of the kernel approximations $ \hat{K}_{\mathrm{LOP}} $ to the original kernel $ K_{\mathrm{LOP}} $. A correction by scaling the parameters $ \hat{\sigma}_k $ with a global factor $ 1 / b_{\mathrm{opt}} $ lowers the error for CLOP~\cite{preiner2014continuous}. Nevertheless, our approximation still better follows $ K_{\mathrm{LOP}} $ with a significantly lower error and is almost unbiased in the $ 1 $\mbox{-}dimensional case.}
    \label{fig:kernel_approximation}
\end{figure}

\begin{table}
    \renewcommand{\arraystretch}{1.1}
    \centering
    \caption{Ratio of Standard Deviations $ \smallnorm{\hat{\mat{\Sigma}}_{\mathrm{LOP}}}^{1 / 2} / \smallnorm{\mat{\Sigma}_{\mathrm{LOP}}}^{1 / 2} $ Between the Kernel Approximations $ \hat{K}_{\mathrm{LOP}} $ and the Original Kernel $ K_{\mathrm{LOP}} $ in $ \mathbb{R}^d $}
    \sisetup{detect-weight=true,mode=text}
    \newrobustcmd{\MakeBold}{\bfseries}
    \begin{tabular}{
        l
        S[table-number-alignment=center, table-format=1.4]
        S[table-number-alignment=center, table-format=1.4]
        S[table-number-alignment=center, table-format=1.4]
        S[table-number-alignment=center, table-format=1.4]
        S[table-number-alignment=center, table-format=1.4]
    }
        \toprule
        & {$ d = 1 $} & {$ d = 2 $} & {$ d = 3 $} & {$ d = 4 $} & {$ d \to \infty $} \\
        \midrule
        CLOP~\cite{preiner2014continuous} & 0.7931 & 0.7632 & 0.7430 & 0.7291 & 0.6659 \\
        Ours & \MakeBold 0.9917 & \MakeBold 0.9834 & \MakeBold 0.9737 & \MakeBold 0.9640 & 0.8881 \\
        Ours (Consistent) & 1.0137 & 1.0252 & 1.0362 & 1.0440 & \MakeBold 1 \\
        \bottomrule
    \end{tabular}
    \label{tab:kernel_standard_deviation}
\end{table}

\begin{figure}
    \centering
    \includegraphics[width=\linewidth, height=0.5\textheight, keepaspectratio]{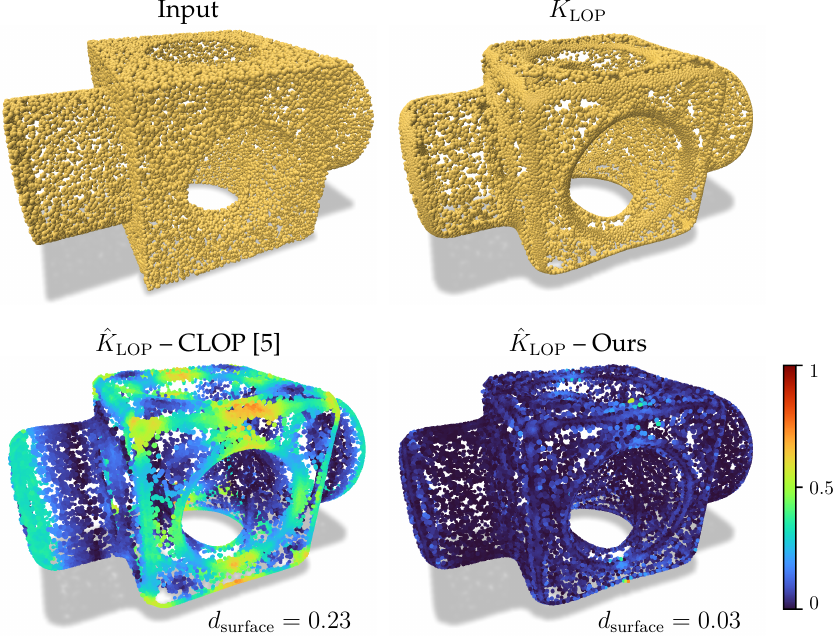}
    \caption{Bias of the kernel approximations $ \hat{K}_{\mathrm{LOP}} $ when applying WLOP~\cite{huang2009consolidation} to smooth the \emph{Block} model (\num{25000} points) with the window size $ { h = 25 } $. Whereas the CLOP~\cite{preiner2014continuous} approximation introduces systematic errors at the edges due to the bias in the width, our variant closely resembles the behavior of the original kernel $ K_{\mathrm{LOP}} $. Unit: $ [\% \text{ BB diagonal}] $.}
    \label{fig:kernel_approximation_comparison}
\end{figure}

In order to evaluate the performance of our density weighting schemes defined in \eqref{eq:simple_scheme} and \eqref{eq:full_scheme} (see Section~\ref{sec:wlop_density_weights}), we measured the regularity of the point cloud $ \set{Q} $ after projection onto a highly irregular target $ \set{P} $~\cite{huang2009consolidation}.
For this purpose, we sampled \num{74000} target points from a 3D surface patch inversely proportional to the intensity of the mapped \emph{Bird} image and took a random subset of \num{3700} points for projection.
Then, we applied \num{100} iterations of the LOP operator as well as its weighted versions and computed the regularity
\begin{equation}
    \sigma_{\set{Q}}
    =
    \left[ \frac{1}{\abs{\set{Q}}} \sum_i \left( d(\vec{q}_i, \set{Q} \tightsetminus \{ \vec{q}_i \}) - \overline{d(\vec{q}_i, \set{Q} \tightsetminus \{ \vec{q}_i \})} \right)^2 \right]^{\frac{1}{2}}
\end{equation}
which is defined as the standard deviation of the nearest neighbor distances $ { d(\vec{x}, \set{Y}) = \min_j \norm{\vec{x} - \vec{y}_j} } $ within the point cloud $ \set{Q} $.
Fig.~\ref{fig:weights_regularity} shows the quantitative results for \num{60}$\times$\num{50} combinations of $ h $ and $ \mu $.
Throughout $ \SI{76.7}{\percent} $ of all combinations, our simple weighting scheme \eqref{eq:simple_scheme} performs better than WLOP with a slightly lower value of $ \sigma_{\set{Q}} $ on average.
Our full scheme \eqref{eq:full_scheme} outperforms WLOP in $ \SI{99.3}{\percent} $ and the simple scheme in $ \SI{98.7}{\percent} $ of all combinations, especially in configurations with low repulsion weights $ { \mu \in [0, 0.2] } $.

\begin{figure*}
    \centering
    \includegraphics[width=\linewidth, height=0.5\textheight, keepaspectratio]{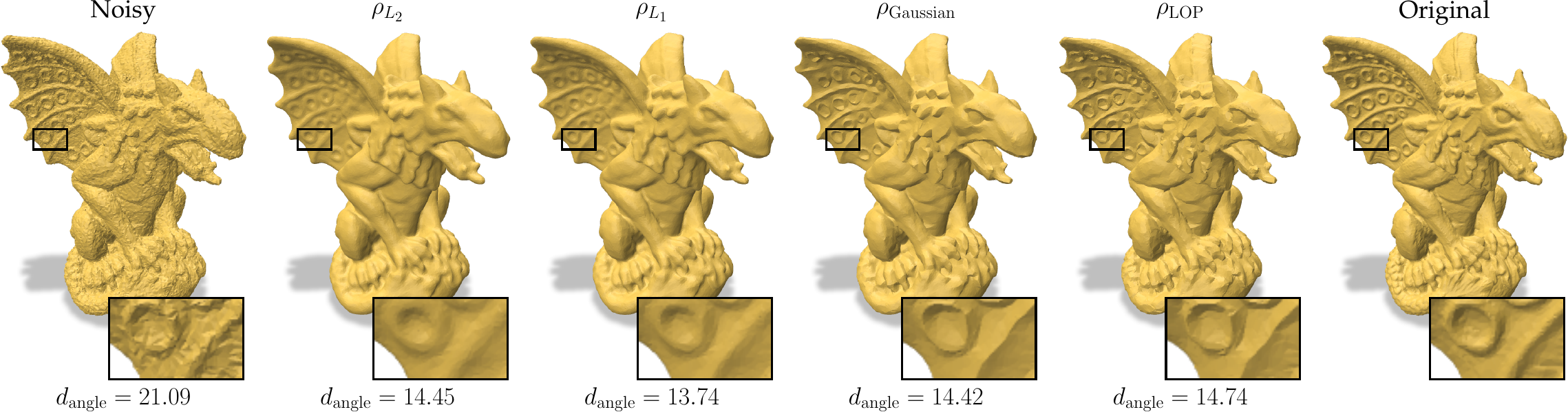}
    \caption{Mesh denoising of the \emph{Gargoyle} model (\num{86311} vertices, \num{172610} faces) corrupted with $ 0.25 \, \overline{l}_e $ uniform noise. Although the mean angular distance $ d_{\mathrm{angle}} $ is slightly higher for the LOP loss $ \rho_{\mathrm{LOP}} $, features and finer details are better preserved. Unit: $ [{\mkern 1mu}^{\circ}] $.}
    \label{fig:normal_filtering_gargoyle}
\end{figure*}

\begin{figure*}
    \centering
    \includegraphics[width=\linewidth, height=0.5\textheight, keepaspectratio]{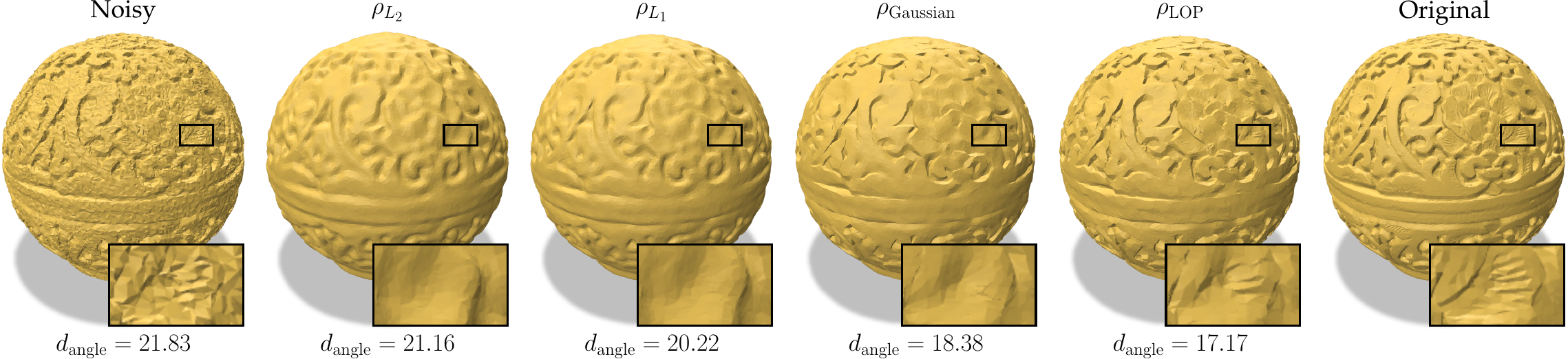}
    \caption{Mesh denoising of the \emph{Box} model (\num{70134} vertices, \num{140259} faces) corrupted with $ 0.25 \, \overline{l}_e $ uniform noise. Filtering with the LOP loss $ \rho_{\mathrm{LOP}} $ results in the lowest mean angular distance $ d_{\mathrm{angle}} $ and reconstructs fine details best. Unit: $ [{\mkern 1mu}^{\circ}] $.}
    \label{fig:normal_filtering_box}
\end{figure*}

These improvements in point cloud regularity directly correspond to a more evenly distributed density along the surface.
Fig.~\ref{fig:weights_density} depicts a comparison of \num{1000}$\times$\num{1000} evenly sampled density values on the respective 3D surface patch.
Both WLOP and our simple scheme \eqref{eq:simple_scheme} normalize the lower frequency components of the density, but still retain high-frequency variations due to the independent computation of each weight.
On the other hand, our full scheme \eqref{eq:full_scheme} does not suffer from these artifacts and only leads to underestimated densities at the boundary and in sparsely sampled regions where the window size $ h $ is not sufficiently large to bridge these gaps.

Since Mean Shift and, thereby, LOP and its variants are scale-invariant with respect to a global normalization constant, we computed the mean density $ \overline{\hat{f}} $ for each weighting scheme and used this value to normalize each density distribution for a fair comparison.
Whereas this value is close to one for both of our schemes due to the correct handling of the normalization constants, we can derive a theoretical estimate of this value for WLOP
\begin{equation}
    \overline{\hat{f}}_{\mathrm{WLOP}}
    \approx
    \frac{1}{\abs{\set{P}} \, h^3} \, c_{\theta}^{(d = 3)} \, \frac{c_{\mathrm{LOP}}^{(d = 3)}}{c_{\mathrm{LOP}}^{(d = 2)}} \frac{c_{\theta}^{(d = 2)}}{c_{\theta}^{(d = 3)}}
    =
    2.396
\end{equation}
which consists of three terms: 1) the normalization constant of the kernel density estimate in \eqref{eq:kernel_density_estimate}; 2) the missing normalization constant of the kernel $ \theta $; and 3) a dimension-dependent correction factor.
The last term models the different domains from which the density is accumulated as we consider a surface patch that corresponds to a 2D subspace embedded in the 3D space.
Therefore, the integration domain of the density differs by one dimension which can be accounted for by the ratio of the normalization constants of both the actual density kernel $ K_{\mathrm{LOP}} $ as well as the chosen kernel $ \theta $ for density weight computation.

\subsection{Evaluation of CLOP Kernel Approximation}
\label{sec:evaluation_of_clop_kernel_approximation}

We evaluated the approximation error of our fitted parameter set in Table~\ref{tab:kernel_approximation} (see Section~\ref{sec:clop_kernel_approximation}) against the original one proposed by CLOP~\cite{preiner2014continuous}.
First, we quantified systematic errors of the kernel approximation $ \hat{K}_{\mathrm{LOP}} $ by analyzing its standard deviation $ \smallnorm{\hat{\mat{\Sigma}}_{\mathrm{LOP}}}^{1 / 2} $ defined by the square root of the magnitude of its covariance matrix \eqref{eq:lop_kernel_approximation_variance}.
Table~\ref{tab:kernel_standard_deviation} indicates that both CLOP and our approximation underestimate the actual value $ \smallnorm{\mat{\Sigma}_{\mathrm{LOP}}}^{1 / 2} $ and that the bias increases in higher dimensions.
Although these errors are significantly lower for our approximation throughout all dimensions, they may still be noticeable.
Our consistent approximation always overestimates the actual standard deviation and has a slightly higher error than the unconstrained variant in low dimensions up to $ { d = 5 } $.
However, it becomes unbiased in the limit $ { d \to \infty } $ and should be preferred in higher dimensions.
We also considered minimizing the $ L_1 $ distance of the kernel approximation $ \hat{K}_{\mathrm{LOP}} $ \eqref{eq:lop_kernel_approximation} to the actual kernel $ K_{\mathrm{LOP}} $ \eqref{eq:lop_kernel} by scaling the values $ \hat{\sigma}_k $ with correction factors $ { 1 / b_{\mathrm{opt}} } $ to obtain an improved set of parameters which is shown in Fig.~\ref{fig:kernel_approximation}.
Here, the error of our approximation is significantly lower than for CLOP both before and after optimal correction.
Furthermore, the optimal scaling factors are similar to the ratios of the standard deviation for $ { d = 1 } $.

In addition to the theoretical analysis of the kernel approximations, we also measured the reconstruction error when replacing the actual kernel in the update step \eqref{eq:lop_update} of the projection with the respective approximation.
For this purpose, we chose the \emph{Block} model and uniformly sampled \num{50000} target points $ \set{P} $ and \num{25000} projection points $ \set{Q} $ respectively.
We applied \num{100} iterations of WLOP as a smoothing operator with a large window size of $ { h = 25 } $ percent of the bounding box diagonal of $ \set{P} $ and a repulsion weight $ { \mu = 0.4 } $.
Then, we measured the distance of each point to the (triangulated) surface of the reference point cloud as well as the mean point-surface distance
\begin{equation}
    d_{\mathrm{surface}}(\set{X}, \set{Y})
    =
    \frac{1}{\abs{\set{X}}} \sum_{i} \min_j d(\vec{x}_i, t(y_j))
    \label{eq:surface_distance}
\end{equation}
where $ t(y_j) $ denotes the $ j $-th triangle of $ \set{Y} $.
Fig.~\ref{fig:kernel_approximation_comparison} shows the results of this point cloud smoothing operation.
Whereas the CLOP approximation introduces higher errors at the edges of the sampled model due to the significantly underestimated standard deviation of the kernel, our approximation does not suffer from these artifacts.

\subsection{Evaluation of Robust Loss Functions}
\label{sec:evaluation_of_robust_loss_functions}

We tested the LOP M\mbox{-}estimator with the corresponding loss $ \rho_{\mathrm{LOP}} $ \eqref{eq:lop_loss} against other popular choices, \ie $ L_2 $, $ L_1 $, and Gaussian M\mbox{-}estimators, for normal filtering in the context of mesh denoising based on the iterative update step in \eqref{eq:normal_update_step} (see Section~\ref{sec:robust_loss_functions}).
For this, we used the \emph{Gargoyle} (\num{86311} vertices, \num{172610} faces) and \emph{Box} (\num{70134} vertices, \num{140259} faces) models and corrupted the vertices in random directions by $ 0.25 \, \overline{l}_e $ uniform noise where $ \overline{l}_e $ denotes the average face edge length.
Then, we applied \num{50} iterations of normal filtering with $ { \sigma = 0.3 } $ for each face normal $ \vec{n} $ within its geometric neighborhood of size $ { r = 1.5 \, \overline{l}_e } $, that is all normals whose face centers are traversable along the surface within a ball of size $ r $.
To avoid the singularity of the $ L_1 $ and LOP losses at $ { x = 0 } $, we only considered the neighboring face normals in the initial iteration and used all normals subsequently.
For the second stage of the mesh denoising framework, we used the vertex update by Zhang \etal~\cite{zhang2018static} with their default parameters of \num{20} iterations and $ { w = 0.001 } $ which avoids the triangle flipping problem.
We evaluated the reconstruction error by the mean angular distance
\begin{equation}
    d_{\mathrm{angle}}(\set{X}, \set{Y})
    =
    \frac{1}{\abs{\set{X}}} \sum_i \left[ \frac{360}{2 \pi} \, \arccos(\dotproduct{\vec{n}(x_i)}{\vec{n}(y_i)}) \right]
\end{equation}
to the face normals of the ground truth mesh.
Fig.~\ref{fig:normal_filtering_gargoyle} and~\ref{fig:normal_filtering_box} show comparisons between the $ L_2 $, $ L_1 $, Gaussian, and LOP loss.
Whereas the $ L_2 $ loss leads to a very smooth surface, its $ L_1 $ counterpart is less sensitive to large normal variations within the local neighborhood and better preserves features.
However, sharp edges cannot be reconstructed since all collected normals are considered in a global fashion.
The Gaussian and LOP loss functions can be viewed as localized versions of the former losses and do not suffer from this limitation.
Finer details being at a similar scale as the applied noise are hard to reconstruct and mostly smoothed out by all variants, but can be partially recovered by the LOP loss.

\subsection{Evaluation of Neural Network Priors}
\label{sec:evaluation_of_neural_network_priors}

We tested the application of different neural network priors for unsupervised point cloud denoising (see Section~\ref{sec:neural_network_priors}).
For this purpose, we used the reference implementation of TotalDenoising~\cite{hermosilla2019total} and trained the method in an unsupervised manner on the real-world large-scale \emph{Paris-rue-Madame} dataset~\cite{serna2014paris}, which was captured by a LiDAR sensor and has no groundtruth data available.
In particular, we employed the same training procedure and default parameter set in all cases, that is the scaling parameter $ { \alpha = 0.5 } $ as well as the standard deviation $ { \sigma = 0.5657 } $ for the kernel in the prior \eqref{eq:network_prior}, and only replaced the Gaussian kernel by the LOP kernel.
Finally, we used the trained models to denoise the test split of the dataset.
Fig.~\ref{fig:total_denoising} shows a comparison between the Gaussian and LOP kernel priors.
While the LiDAR noise is reliably removed from the test datasets for both kernels, filtering with the Gaussian kernel leads to significant oversmoothing of smaller features and, in addition, to a noticable shrinkage of the point cloud.
Furthermore, holes may be introduced at sharp features like edges since the Gaussian prior tends to move nearby points into clusters apart from these features.
In contrast to this, the LOP kernel with equal standard deviation $ \sigma $ preserves higher frequency information and, in turn, sharp features without introducing holes.
The shrinkage effect also is greatly mitigated which becomes apparent at thin structures, like the sash bars in the windows, where the width is significantly better preserved and closely matches the expected width from the input data.

\begin{figure}
    \centering
    \includegraphics[width=\linewidth, height=0.8\textheight, keepaspectratio]{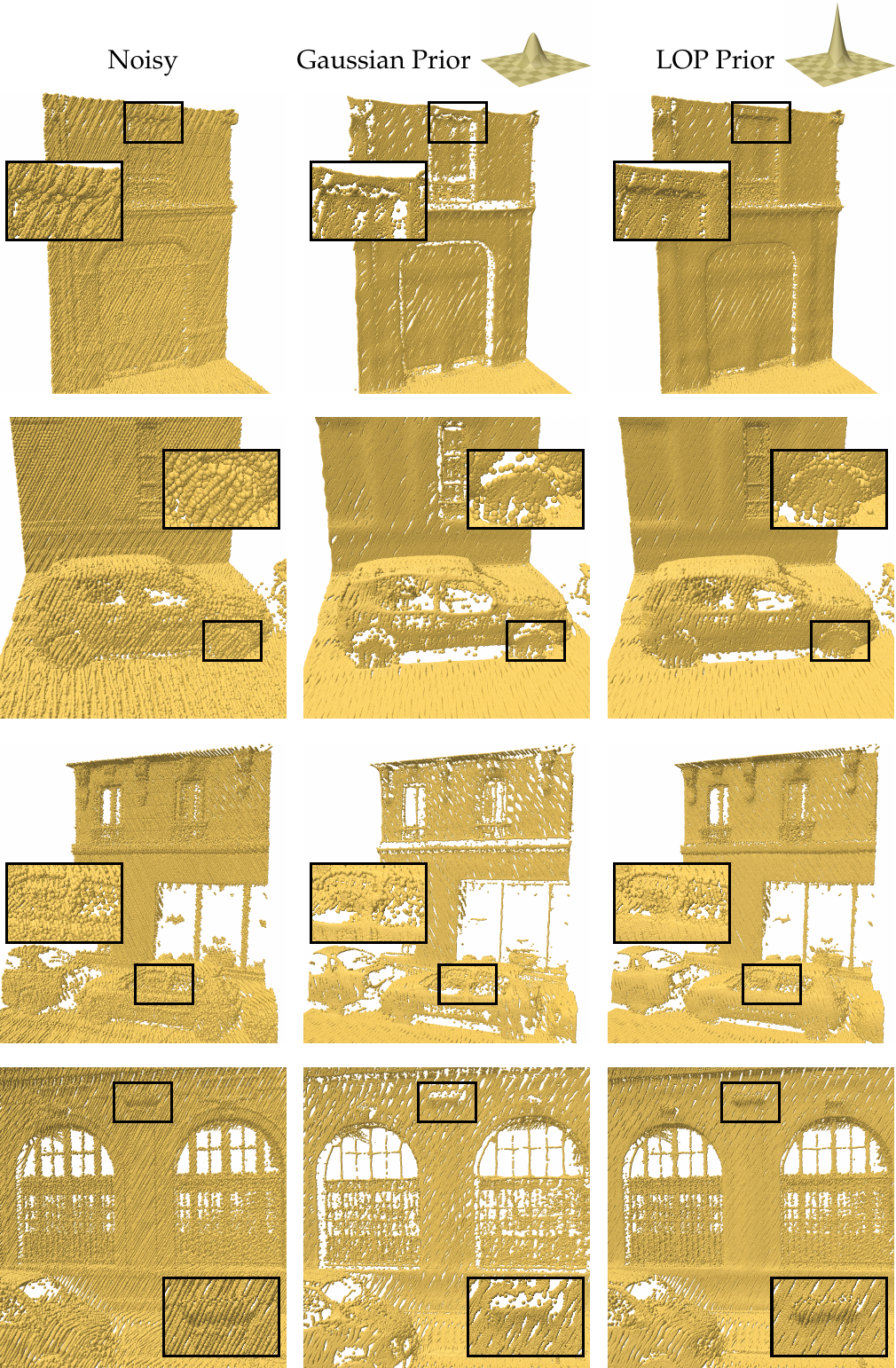}
    \caption{Point cloud denoising of the \emph{Paris-rue-Madame} dataset with TotalDenoising~\cite{hermosilla2019total} for the Gaussian and LOP kernels as neural network priors. Performing the unsupervised training with the LOP prior better preserves features and edges and leads to significantly less oversmoothing and shrinkage. Furthermore, using the LOP prior also better handles the systematic pattern induced by the scanning process.}
    \label{fig:total_denoising}
\end{figure}

\begin{table}
    \renewcommand{\arraystretch}{1.2}
    \setlength{\tabcolsep}{4.8pt}
    \centering
    \caption{Comparison Between Unsupervised Learning-based Point Cloud Denoising Methods. Unit: $ [10^{-4}] $}
    \sisetup{detect-weight=true,mode=text}
    \newrobustcmd{\MakeBold}{\bfseries}
    \begin{tabular}{
        l
        S[table-number-alignment=center, table-format=1.3]
        S[table-number-alignment=center, table-format=1.3]
        S[table-number-alignment=center, table-format=1.3]
        S[table-number-alignment=center, table-format=1.3]
        S[table-number-alignment=center, table-format=1.3]
        S[table-number-alignment=center, table-format=1.3]
    }
        \toprule
        & \multicolumn{2}{c}{$ \sigma_{\mathrm{noise}} = 0.5 $} & \multicolumn{2}{c}{$ \sigma_{\mathrm{noise}} = 1 $} & \multicolumn{2}{c}{$ \sigma_{\mathrm{noise}} = 1.5 $} \\
        \cmidrule(lr){2-3} \cmidrule(lr){4-5} \cmidrule(lr){6-7}
        & {$ d_{\mathrm{CD}} $} & {$ d_{\mathrm{P2M}} $} & {$ d_{\mathrm{CD}} $} & {$ d_{\mathrm{P2M}} $} & {$ d_{\mathrm{CD}} $} & {$ d_{\mathrm{P2M}} $} \\
        \midrule
        Score (Unsup.)~\cite{luo2021score} & \MakeBold 0.150 & \MakeBold 0.115 & 0.404 & 0.315 & 0.821 & 0.609 \\
        TD ($ K_{\mathrm{Gaussian}} $)~\cite{hermosilla2019total} & 0.173 & 0.150 & 0.343 & 0.303 & 0.466 & 0.428 \\
        TD ($ K_{\mathrm{LOP}} $) & 0.175 & 0.148 & \MakeBold 0.282 & \MakeBold 0.246 & \MakeBold 0.424 & \MakeBold 0.389 \\
        \bottomrule
    \end{tabular}
    \label{tab:synthetic_denoising_comparison}
\end{table}

In addition to these qualitative results, we also performed a quantitative comparison between recent learning-based approaches.
For a fair comparison, we only considered unsupervised methods, that is TotalDenoising (TD)~\cite{hermosilla2019total} with the original Gaussian kernel prior as well as with our LOP kernel prior, and Score-based denoising~\cite{luo2021score} which, although being supervised, also supports training in an unsupervised manner.
We used the \emph{ModelNet40} dataset~\cite{wu20153d} provided by TD~\cite{hermosilla2019total} that consists of 15 collected object classes with 5 training meshes and 2 test meshes per class, from which point clouds were sampled using Poisson disk sampling and corrupted by Gaussian noise with standard deviations of $ { \sigma_{\mathrm{noise}} = 0.5, 1, 1.5 } $ percent of the bounding box diagonal.
After training all methods with default parameters on this dataset, we applied them to denoise the test data using $ 1, 4, 7 $ iterations with TD and $ 1, 1, 2 $ iterations with Score-based denoising respectively for the different noise levels.
We measured the performance with the commonly used Chamfer distance
\begin{equation}
    d_{\mathrm{CD}}(\set{X}, \set{Y})
    =
    \frac{1}{\abs{\set{X}}} \sum_{i} d(\vec{x}_i, \set{Y})^2
    +
    \frac{1}{\abs{\set{Y}}} \sum_{i} d(\vec{y}_i, \set{X})^2
\end{equation}
where $ { d(\vec{x}_i, \set{Y}) = \min_j \norm{\vec{x}_i - \vec{y}_j} } $ denotes the nearest neighbor distance of $ \vec{x}_i $ to the point cloud $ \set{Y} $ as well as with the point-to-mesh distance
\begin{equation}
\begin{split}
    d_{\mathrm{P2M}}(\set{X}, \set{Y})
    = \ &
    \frac{1}{\abs{\set{X}}} \sum_{i} \min_j d(\vec{x}_i, t(y_j))^2
    \\
    & +
    \frac{1}{\abs{\set{Y}}} \sum_{j} \min_i d(\vec{x}_i, t(y_j))^2
\end{split}
\end{equation}
where $ d(\vec{x}_i, t(y_j)) $ is the distance of $ \vec{x}_i $ to the triangle $ t(y_j) $ as in \eqref{eq:surface_distance}.
Table~\ref{tab:synthetic_denoising_comparison} shows the results of all methods averaged over the test data.
At the low noise level $ { \sigma_{\mathrm{noise}} = 0.5 } $, Score-based denoising achieves slightly better results than TotalDenoising where the performance is very similar between the Gaussian and the LOP kernel prior.
On the other hand, applying the LOP kernel prior significantly increases the robustness of TotalDenoising at higher noise levels and also outperforms all other approaches with respect to both the Chamfer and the point-to-mesh distance.

%-------------------------------------------------------------------------
\section{Conclusions}

We presented incomplete gamma kernels, a novel family of kernels generalizing LOP operators.
By revisiting the classical localized $ L_1 $ estimator used in LOP, we revealed its relation to the Mean Shift framework via a novel kernel $ K_{\mathrm{LOP}} $ and generalized this result to arbitrary localized $ L_p $ estimators.
We derived several theoretical properties of the kernel family $ K_{\Gamma} $ concerning distributional, Mean Shift induced, and other aspects such as strict positive definiteness to obtain a deeper understanding of the operator's projection behavior.
Furthermore, we illustrated several applications including an improved WLOP density weighting scheme, a more accurate kernel approximation for CLOP, incomplete gamma losses $ \rho_{\Gamma} $ as a novel set of robust loss functions, as well as better neural network priors and confirmed their effectiveness in a variety of quantitative and qualitative experiments.
We expect that building upon the insights provided by our work will be beneficial for future developments on point cloud denoising as well as in many other related fields, including applications beyond pure denoising.

%-------------------------------------------------------------------------

%-------------------------------------------------------------------------
\appendix[Integral of Incomplete Gamma and Bessel Functions]

Let $ { a > 0, b > 0, s \ge 0 } $ be real-valued constants.
Then for any real-valued $ { p > -1 } $, the relation
\begin{equation}
\begin{split}
    & \int_{0}^{\infty} \Gamma(s, a^2 \, x^2) \, \bessel_p(b \, x) \, x^{p + 1} \, \mathrm{d} x
    \\
    = \ &
    \frac{1}{2} \, \left( \frac{b}{2} \right)^p \frac{1}{a^{2 p + 2}} \frac{\Gamma(p + s + 1)}{\Gamma(p + 2)} \, { \textstyle \hyp1f1( p + s + 1, p + 2, - \frac{b^2}{4 a^2} ) }
\end{split}
\end{equation}
holds where $ \Gamma(a, x) $ denotes the upper incomplete gamma function, $ \bessel_p $ the Bessel function of order $ p $, and $ \hyp1f1 $ the confluent hypergeometric function of the first kind.

\begin{IEEEproof}
First, we plug in the definition of the incomplete gamma function and change the order of integration:
\begin{equation}
\begin{split}
    & \int_{0}^{\infty} \Gamma(s, a^2 \, x^2) \, \bessel_p(b \, x) \, x^{p + 1} \, \mathrm{d} x
    \\
    = \ &
    \int_{0}^{\infty} \int_{a^2 x^2}^{\infty} t^{s - 1} \, \euler^{- t} \, \bessel_p(b \, x) \, x^{p + 1} \, \mathrm{d} t \, \mathrm{d} x
    \\
    = \ &
    \int_{0}^{\infty} \int_{0}^{\frac{1}{a} \sqrt{t}} t^{s - 1} \, \euler^{- t} \, \bessel_p(b \, x) \, x^{p + 1} \, \mathrm{d} x \, \mathrm{d} t
    \\
    = \ &
    \int_{0}^{\infty} t^{s - 1} \, \euler^{- t} \, \left[ \int_{0}^{\frac{1}{a} \sqrt{t}} \bessel_p(b \, x) \, x^{p + 1} \, \mathrm{d} x \right] \, \mathrm{d} t
\end{split}
\end{equation}
The inner integral can be solved by substituting $ { y = b \, x } $ and then applying the well-known relation $ { \int x^{p + 1} \, \bessel_{p}(x) \, \mathrm{d} x = x^{p + 1} \, \bessel_{p + 1}(x) } $:
\begin{equation}
\begin{split}
    & \int_{0}^{\infty} t^{s - 1} \, \euler^{- t} \, \left[ \int_{0}^{\frac{1}{a} \sqrt{t}} \bessel_p(b \, x) \, x^{p + 1} \, \mathrm{d} x \right] \, \mathrm{d} t
    \\
    = \ &
    \int_{0}^{\infty} t^{s - 1} \, \euler^{- t} \, \left[ \frac{1}{b^{p + 2}} \int_{0}^{\frac{b}{a} \sqrt{t}} \bessel_p(y) \, y^{p + 1} \, \mathrm{d} y \right] \, \mathrm{d} t
    \\
    = \ &
    \int_{0}^{\infty} t^{s - 1} \, \euler^{- t} \, \frac{1}{b^{p + 2}} \, { \textstyle \bessel_{p + 1}(\frac{b}{a} \sqrt{t}) } \, \left( \frac{b}{a} \sqrt{t} \right)^{p + 1} \, \mathrm{d} t
    \\
    = \ &
    \frac{1}{b} \frac{1}{a^{p + 1}} \int_{0}^{\infty} \left( \sqrt{t} \right)^{p + 2 s - 1} \, \euler^{- t} \, { \textstyle \bessel_{p + 1}(\frac{b}{a} \sqrt{t}) } \, \mathrm{d} t
\end{split}
\end{equation}
We can now substitute $ { u = \sqrt{t} } $ and obtain a closed-form solution of the remaining integral~\cite{bateman1953higher} which holds under the condition $ { 2 p + 2 s + 2 > 0 } $:
\begin{equation}
\begin{split}
    & \frac{1}{b} \frac{1}{a^{p + 1}} \int_{0}^{\infty} \left( \sqrt{t} \right)^{p + 2 s - 1} \, \euler^{- t} \, { \textstyle \bessel_{p + 1}(\frac{b}{a} \sqrt{t}) } \, \mathrm{d} t
    \\
    = \ &
    \frac{1}{b} \frac{1}{a^{p + 1}} \, 2 \int_{0}^{\infty} u^{p + 2 s} \, \euler^{- u^2} \, { \textstyle \bessel_{p + 1}(\frac{b}{a} u) } \, \mathrm{d} u
    \\
    = \ &
    \frac{1}{b} \frac{1}{a^{p + 1}} \, 2 \, \Bigg[ \frac{1}{2} \frac{\Gamma(p + s + 1)}{\Gamma(p + 2)} \, \left( \frac{1}{2} \frac{b}{a} \right)^{p + 1}
    \\
    &
    \phantom{\frac{1}{b} \frac{1}{a^{p + 1}} \, 2} \, \quad { \textstyle \hyp1f1( p + s + 1, p + 2, - \frac{b^2}{4 a^2} ) } \Bigg]
    \\
    = \ &
    \frac{1}{2} \, \left( \frac{b}{2} \right)^p \frac{1}{a^{2 p + 2}} \frac{\Gamma(p + s + 1)}{\Gamma(p + 2)} \, { \textstyle \hyp1f1( p + s + 1, p + 2, - \frac{b^2}{4 a^2} ) }
\end{split}
\end{equation}
Considering the assumption $ { s \ge 0 } $, this condition simplifies to $ { p > -1 } $.
\end{IEEEproof}

A similar, more restricted result for the special case $ { s = 1 / 2 } $ has also been shown in previous work~\cite{ng1969table}.

%-------------------------------------------------------------------------

\ifCLASSOPTIONcompsoc
  \section*{Acknowledgments}
\else
  \section*{Acknowledgment}
\fi

This work was supported by the DFG projects KL 1142/11-2 (DFG Research Unit FOR 2535 Anticipating Human Behavior) and NFDI4Culture (DFG Project Number 441958017).
We thank the reviewers for their valuable comments which helped us to improve the quality of our manuscript.

\ifCLASSOPTIONcaptionsoff
  \newpage
\fi

%\IEEEtriggeratref{69}


\bibliographystyle{IEEEtran}
\bibliography{tex/literature}

\end{document}